\documentclass[screen,sigconf]{acmart}

\usepackage{graphicx}
\usepackage{booktabs}
\usepackage{array}
\usepackage{multirow}
\usepackage{multicol}
\usepackage{float}
\usepackage{dblfloatfix}
\usepackage{xcolor}
\usepackage{rotating}
\usepackage{subcaption}
\usepackage{longtable}
\usepackage{listings}
\usepackage{lscape}
\newcolumntype{P}[1]{>{\centering\arraybackslash}p{#1}}

\AtBeginDocument{%
  \providecommand\BibTeX{{%
    \normalfont B\kern-0.5em{\scshape i\kern-0.25em b}\kern-0.8em\TeX}}}
    \def\LT@makecaption#1#2#3{%
    \multicolumn{\LT@cols}{|c|}{\cellcolor{gray!40}\textbf{#2: }\textbf{#3}}
}

\copyrightyear{2023} 
\acmYear{2023} 
\setcopyright{rightsretained} \acmConference[AIES '23]{AAAI/ACM Conference on AI, Ethics, and Society}{August 8--10, 2023}{Montréal, QC, Canada} 
\acmBooktitle{AAAI/ACM Conference on AI, Ethics, and Society (AIES '23), August 8--10, 2023, Montréal, QC, Canada} \acmDOI{10.1145/3600211.3604719} 
\acmISBN{979-8-4007-0231-0/23/08}
\begin{document}
\title[Evaluating the Impact of Social Determinants on Health Prediction in the Intensive Care Unit]{Evaluating the Impact of Social Determinants \\ on Health Prediction in the Intensive Care Unit}

\author{Ming Ying Yang}
\email{ming1022@mit.edu}
\affiliation{
  \institution{Massachusetts Institute of Technology}
  \country{USA}
}

\author{Gloria Hyunjung Kwak}
\email{hkwak1@mgh.harvard.edu}
\affiliation{
  \institution{Harvard Medical School}
  \institution{Massachusetts General Hospital}
  \country{USA}
}

\author{Tom Pollard}
\email{tpollard@mit.edu}
\affiliation{
  \institution{Massachusetts Institute of Technology}
  \country{USA}
}

\author{Leo Anthony Celi}
\email{lceli@mit.edu}
\affiliation{
  \institution{Massachusetts Institute of Technology} 
  \institution{Beth Israel Deaconess Medical Center} 
  \country{USA}
}
\author{Marzyeh Ghassemi}
\email{mghassem@mit.edu}
\affiliation{
  \institution{Massachusetts Institute of Technology}
  \country{USA}
}

\renewcommand{\shortauthors}{Yang et al.}

\begin{abstract}

Social determinants of health (SDOH) -- the conditions in which people live, grow, and age -- play a crucial role in a person's health and well-being. There is a large, compelling body of evidence in population health studies showing that a wide range of SDOH is strongly correlated with health outcomes. Yet, a majority of the risk prediction models based on electronic health records (EHR) do not incorporate a comprehensive set of SDOH features as they are often noisy or simply unavailable. 
Our work links a publicly available EHR database, MIMIC-IV, to well-documented SDOH features. We investigate the impact of such features on common EHR prediction tasks across different patient populations.
We find that community-level SDOH features do not improve model performance for a general patient population, but can improve data-limited model fairness for specific subpopulations. We also demonstrate that SDOH features are vital for conducting thorough audits of algorithmic biases beyond protective attributes. We hope the new integrated EHR-SDOH database will enable studies on the relationship between community health and individual outcomes and provide new benchmarks to study algorithmic biases beyond race, gender, and age. 


\end{abstract}





\maketitle

\section{Introduction}
The increasing adoption of electronic health records (EHRs) in modern healthcare systems has facilitated the development of machine learning (ML) models to predict the progression of diseases and patient outcomes. Many such models \cite{lehman_risk_2012, ghassemi_unfolding_2014,harutyunyan_multitask_2019} incorporate clinical factors (e.g., labs, vitals, medication, procedures) and basic demographic features (e.g., age, gender, and race) to identify high-risk patients. However, a patient's clinical profile only offers a partial view of all the risk factors that affect their health. Understanding the conditions of their living environment may help to fill in the missing pieces and benefit patients' health and medical care. Human health is affected by many non-clinical factors, commonly known as social determinants of health (SDOH). 

The Healthy People 2030 initiative \cite{healthy_people_2030_social_2020}, developed by the US Department of Health and Human Services, describes SDOH as "the conditions in the environments where people are born, live, learn, work, play, worship, and age that affect a wide range of health, functioning, and quality-of-life outcomes and risks." They grouped SDOH into five key domains: (1) economic stability \cite{shain_health_2004, brooker_gender_2001}, (2) education access and quality \cite{hahn_education_2015, krueger_mortality_2015}, (3) health care and quality \cite{hood_county_2016}, (4) neighborhood and built environment \cite{kruger_neighborhood_2007, mitchell_current_2007}, and (5) social and community context \cite{chang_social_2019, meddings_impact_2017}. 

Across all five domains, SDOH can have either a direct or indirect impact on one's health. At a high level, they can be viewed as individual-level determinants or community-level determinants \cite{cantor_integrating_2018}. The former determinants are specific to a person, and examples include education level, annual income, and family dynamics. Access to individual-level SDOH is limited due to the lack of standardized and validated SDOH screening questions \cite{cantor_integrating_2018} and privacy concerns \cite{mcgraw_privacy_2015}. In contrast, community-level SDOH measure broader socioeconomic, neighborhood, and environmental characteristics such as unemployment rate, access to public transportation, and air pollution levels. They serve as ``community vital sign'' \cite{bazemore_community_2016} that reflect complex societal factors and health disparities that influence one's health \cite{malina_race_2021,adler_addressing_2016}.

Population health studies have identified many SDOH to be strongly correlated with acute and chronic conditions \cite{de_boer_predictors_1997,amrollahi_inclusion_2022,pfohl_creating_2019,galobardes_systematic_2006,dugravot_social_2020}. SDOH are also underlying, contributing factors of health disparities (e.g., poverty \cite{garcia_popa-lisseanu_determinants_2005, cooper_poverty_2012, vutien_utilization_2019}, unequal access to health care \cite{chang_social_2019, cristancho_listening_2008}, low educational attainment \cite{assari_education_2018, cutler_education_2006, hahn_education_2015}, and segregation \cite{prrac_racial_2017, carter_addressing_2022}).
However, to date, there has been less focus in the ML community to include SDOH in common EHR prediction tasks because many SDOH measures are poorly collected, lack granularity, or are simply unavailable. An American Health Information Management Association (AHIMA) survey \cite{norc_at_the_university_of_chicago_social_2023} finds that most healthcare organizations are collecting SDOH data, but they face challenges with a lack of standardization and integration of the SDOH data into EHR and patient distrust in sharing the data. Thus, while SDOH are being increasingly studied in population health \cite{hood_county_2016, yu_social_2020, purtle_uses_2019} and primary care settings \cite{katz_association_2018, norc_at_the_university_of_chicago_social_2023}, data limitations have left the association between SDOH and critical care outcomes largely unexplored.

In this work, we investigate the impact of incorporating SDOH features on common EHR prediction tasks in the intensive care unit (ICU). We first link MIMIC-IV \cite{johnson_alistair_mimic-iv_2022}, a publicly available EHR database, to external SDOH databases based on patient zip code. We then train models on the common tasks of mortality and readmission risk, evaluating the contribution of SDOH as compared to the EHR data alone. We find that adding SDOH does not improve model performance in the general patient population. We do note that, as compared to the EHR data alone, incorporating SDOH can lead to better-calibrated and fairer models in specific subgroups, with varying levels of contribution depending on the population and predictive task. Finally, we illustrate that fairness audits based on both protective attributes and SDOH features help to connect the commonly observed disparities to the underlying mechanisms that drive adverse health outcomes downstream.

Our work makes three main contributions. 
\begin{itemize}
    \item We release a publicly accessible database that combines EHR data with SDOH measures. To the best of our knowledge, this is the first public EHR database that contains structural features that span all five defined SDOH domains. The database will enable new studies on the relationship between community health and individual clinical outcomes. 
    \item We investigate the impact of incorporating SDOH in predictive models across three tasks, three model classes, and six patient populations. We find that the inclusion of SDOH can improve performance for certain vulnerable subgroups.
    \item We demonstrate that SDOH features enable more fine-grain audits of algorithmic fairness, reporting the FPR parity -- the difference in false positive rates (FPR) -- across intersectional patient subgroups. 
\end{itemize}

\section{Related Work}
\subsection{SDOH in Health Prediction}
A number of studies in population health have attempted to assess the impact of social factors on health \cite{project_social_2003, marmot_social_2005, braveman_social_2014, adler_addressing_2016}. There is a large, compelling body of evidence showing that a wide range of SDOH is strongly correlated with health outcomes, such as sepsis \cite{amrollahi_inclusion_2022}, heart failure \cite{chin_correlates_1997}, pneumonia \cite{meddings_impact_2017}, cardiovascular disease \cite{galobardes_systematic_2006}, and diabetes \cite{ye_predicting_2020, walker_relationship_2014}. A particular study found that 40\% of deaths in the United States are caused by behavior patterns that could be modified by preventive interventions and suggested that only 10-15\% of preventable mortality could be avoided by higher-quality medical care \cite{mcginnis_actual_1993}. Other studies have also indicated that the effect of medical care may be more limited than commonly believed \cite{hood_county_2016, mackenbach_contribution_1989, mackenbach_contribution_1996}. However, there are active controversies regarding the strength of the evidence that suggests a causal relationship between SDOH and well-being. These researchers are increasingly utilizing SDOH to predict individual health outcomes \citep{shadmi_predicting_2015, ozyilmaz_worse_2022}.

While several studies have shown that machine learning models can predict individual patient outcomes, such as in-hospital mortality \citep{ghassemi_unfolding_2014, grnarova_neural_2016, lehman_risk_2012, ye_predicting_2020, moon_survlatent_2022} and readmission \citep{lehman_risk_2012, golmaei_deepnote-gnn_2021, rumshisky_predicting_2016}, very few have incorporated SDOH into the models due to the lack of granular and high-quality SDOH data at the individual level.

Due to the limited availability of individual-level SDOH data, many studies are limited to community-level SDOH data \citep{chen_social_2020}. Most found that community-level SDOH do not lead to improvement in model performance \citep{bhavsar_value_2018, jamei_predicting_2017,vest_prediction_2019}, partly due to the low data resolution. In contrast, researchers who are able to access individual-level SDOH generally report improvements in the model's predictive performance \citep{chen_social_2020, feller_using_2018, molfenter_roles_2012}. These studies often focus on a specific outcome for a specific patient group, such as HIV risk assessment \cite{feller_using_2018, nijhawan_clinical_2019} and suicide attempts \cite{walsh_predicting_2018, zheng_development_2020}. There have also been studies of model improvements for readmission and mortality prediction in specific subgroups, such as the elderly and obese \cite{zhang2020assessing}. One has shown that integrating SDOH into predictive models can improve the fairness of the prediction in underserved heart failure subpopulations \cite{li_improving_2022}.

Despite a growing body of SDOH-focused research, the relationship between SDOH and critical care outcomes is unclear. While some have argued that the ICU is not an appropriate setting to collect and identify SDOH, there are several reasons why it could be essential. For example, critical conditions place high demands on the patient and their social network \cite{mcpeake_modification_2022,turnbull_perceived_2022}. Social isolation may increase the risk of adverse outcomes, such as mortality \cite{turnbull_perceived_2022}. By incorporating SDOH into MIMIC-IV, our work investigates the contribution of community-level SDOH on common EHR prediction tasks in a multi-year cohort in the critical care setting.

\subsection{SDOH in EHR}
In order for SDOH features to be readily incorporated into risk prediction models, they need to be collected and documented with individual health outcomes. EHRs contain clinical information about patients, such as medical history, vital signs, laboratory data, immunizations, and medications \citep{centers_for_medicare__medicaid_services_electronic_2021}. In the United States, few SDOH features are currently documented in structured EHR data fields due to the lack of adoption of standardized and validated SDOH screening questions \cite{cantor_integrating_2018}. The set of SDOH features available for research use is typically limited to insurance type, preferred language, and smoking and alcohol use, but SDOH information can also be extracted from unstructured EHR data (i.e. clinical notes) \citep{navathe_hospital_2018,brown_information_2022,feller_using_2018}. SDOH may also be captured in billing codes \cite{american_hospital_association_icd-10-cm_2022}, but they have not been widely utilized by providers \cite{jessica_l_maksut_utilization_2021}. 

The integration of SDOH in EHRs is further delayed due to concerns regarding privacy and misuse \cite{mcgraw_privacy_2015}. The United States Public Health Services Syphilis Study at Tuskegee (Tuskegee Study) among African Americans \cite{washington_medical_2006} and efforts to sterilize American Indians \cite{lawrence_indian_2000, davis_practicing_1999} are examples of a dark history of structural inequities in healthcare and unethical medical experimentation against racial and ethnic minorities. As a result, mistrust of the healthcare system and medical research has been well documented among minority groups \cite{boag_racial_2018, dula_african_1994, corbie-smith_distrust_2002}. The collection and utilization of SDOH require the consent and trust of the patients. Patients who identify with populations that medical establishments and medical researchers historically mistreat might not want to share any personal or sensitive information.

Overall, current EHR-derived SDOH data do not constitute a comprehensive set of SDOH domains. In this study, we link a large, multi-year EHR dataset to public SDOH datasets covering \emph{all five SDOH domains} to comprehensively study the relationship between the community-level SDOH and patient outcomes.

\subsection{Fairness and Bias in Healthcare}
While much work has been done in algorithmic fairness and bias in health, most of the studies that focused on group-based fairness have only examined bias from the lens of protected attributes, namely age, gender, and race \cite{chen_can_2019, chen_why_2018, hashimoto_fairness_2018,obermeyer_dissecting_2019, seyyed-kalantari_chexclusion_2020,balagopalan_road_2022,zhang_hurtful_2020,seyyed-kalantari_underdiagnosis_2021, zhang_improving_2022,suriyakumar_when_2023, adam_write_2022, hebert_methods_2017,maness_social_2021}. Recent fairness literature has underscored the importance of measuring biases from multidimensional perspectives, focusing on social processes that produce the biases \cite{hanna_towards_2020, selbst_fairness_2018, hoffmann_where_2019}. 

There is strong evidence that intersectional social identities are related to a patient’s health outcomes \cite{katz_association_2018, seyyed-kalantari_underdiagnosis_2021}. Capturing social context beyond protected attributes in the form of SDOH is thus vital for this cause. For example, in the primary care setting, researchers have observed a negative correlation between the odds of receiving appropriate prevention and screening and the number of social risk factors experienced by the patient \cite{katz_association_2018}. The more factors a patient was living with, the less likely they were to receive care such as a mammogram screening or vaccinations. This is not something that can be detected through race or gender alone. 

Moreover, a recent meta-analysis \cite{vince_evaluation_2023} ranked 47 studies using a self-developed SDOH scoring system based on the type and number of SDOH features used. The researchers found that Black patients had significantly higher prostate cancer–specific mortality (PCSM) than White patients when there was minimal accounting for SDOH. In contrast, studies with greater consideration for SDOH showed the opposite: Black patients had significantly lower PCSM compared to White patients. The findings of this meta-analysis should not be interpreted as suggesting that racial disparities do not exist. Rather, it suggests that there is a significant interaction between race and SDOH, and health equity researchers should incorporate SDOH features into data collection and analyses to better address the long-standing disparities in healthcare \cite{washington_medical_2006}. 

We hope the new integrated MIMIC-IV-SDOH dataset will enable more studies that follow the complex hierarchical system that defined advantaged or disadvantaged subjects in the first place. Our work serves as a first effort, and we demonstrate how SDOH features allow for more granular, actionable algorithmic audits. 

\section{EHR-SDOH Database: MIMIC-IV-SDOH}
The MIT Laboratory for Computational Physiology (LCP) developed and maintains the publicly available Medical Information Mart for Intensive Care (MIMIC), a database on patients admitted to the emergency department (ED) and intensive care units (ICU) at the Beth Israel Deaconess Medical Center (BIDMC) in Boston \cite{johnson_mimic_2018}. The database is used by researchers in over 30 countries for clinical research studies, exploratory analyses, and the development of decision-support tools \cite{pollard_enabling_2017}. The current version, MIMIC-IV, contains detailed, de-identified data associated with over 70,000 ICU stays from 2008 to 2019 and over 400,000 ED stays from 2011 and 2019. Yet, due to the lack of high-quality SDOH data, none of the studies or tools built based on MIMIC account for SDOH measures beyond basic demographics such as insurance, and language. To enable the study of the relationship between community characteristics and individual health outcomes, we create the MIMIC-IV-SDOH database by linking MIMIC-IV to three public SDOH databases (\autoref{tab:sdoh_data}):
 \begin{enumerate}
     \item County Health Rankings (CHR) \cite{county_health_rankings__roadmaps_massachusetts_2022}
     \item Social Vulnerability Index (SVI) \cite{cdcatsdr_cdcatsdr_2022}
     \item Social Determinants of Health Database (SDOHD) \cite{agency_for_healthcare_research_and_quality_social_2022}
 \end{enumerate} 
This database will be made available on PhysioNet \cite{goldberger_physiobank_2000}.

\begin{figure*}[!b]
    \centering
    \begin{subfigure}[b]{.57\textwidth}
        \includegraphics[width=1\textwidth]{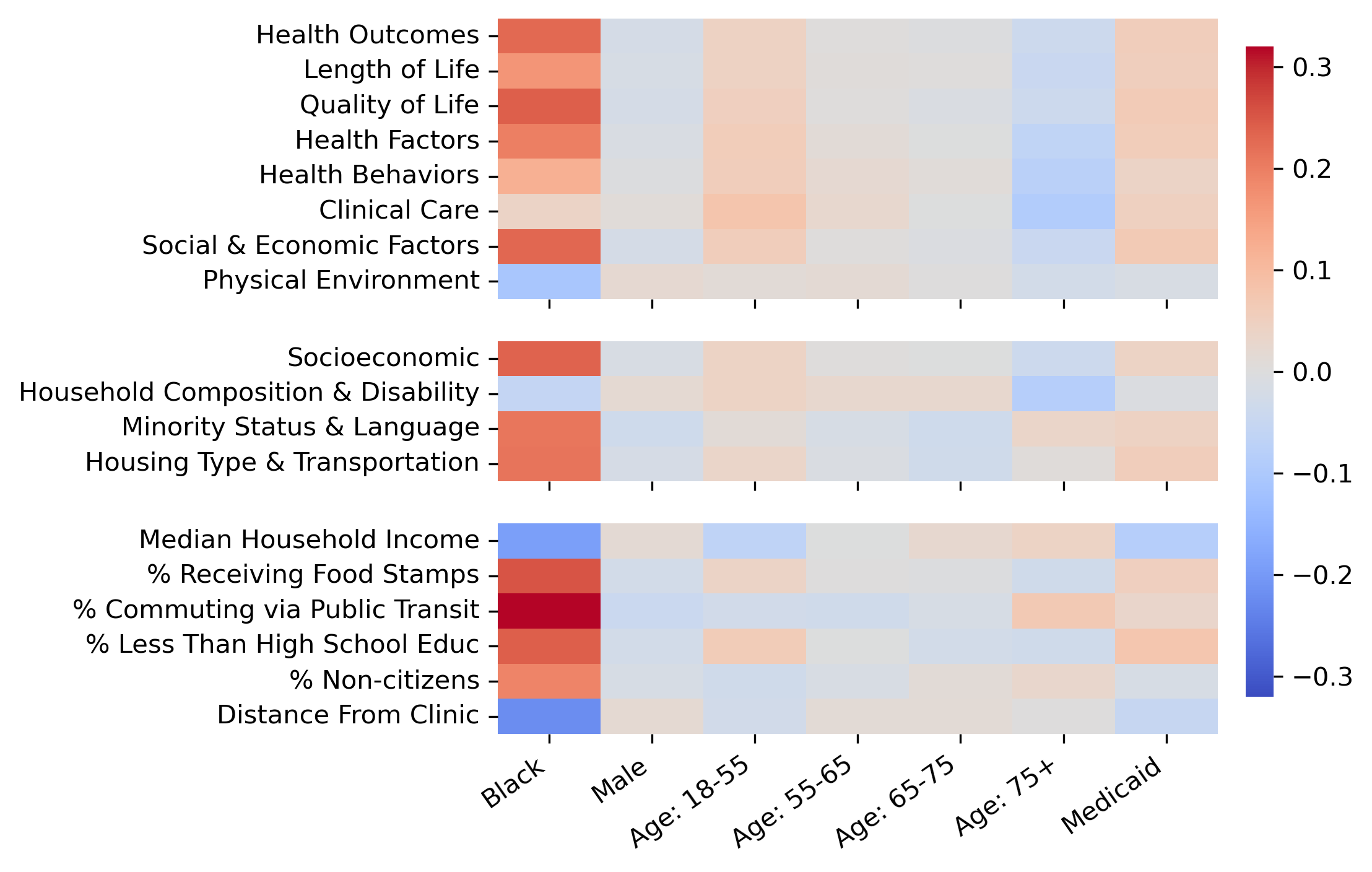}
        \caption{Pearson correlation coefficients between basic demographic features and selected features in CHR (top), SVI (middle), and SDOHD (bottom). }
        \label{fig:heatmap}
    \end{subfigure}
    \hspace*{1cm}
    \begin{subfigure}[b]{.33\textwidth}
        \includegraphics[width=1\textwidth]{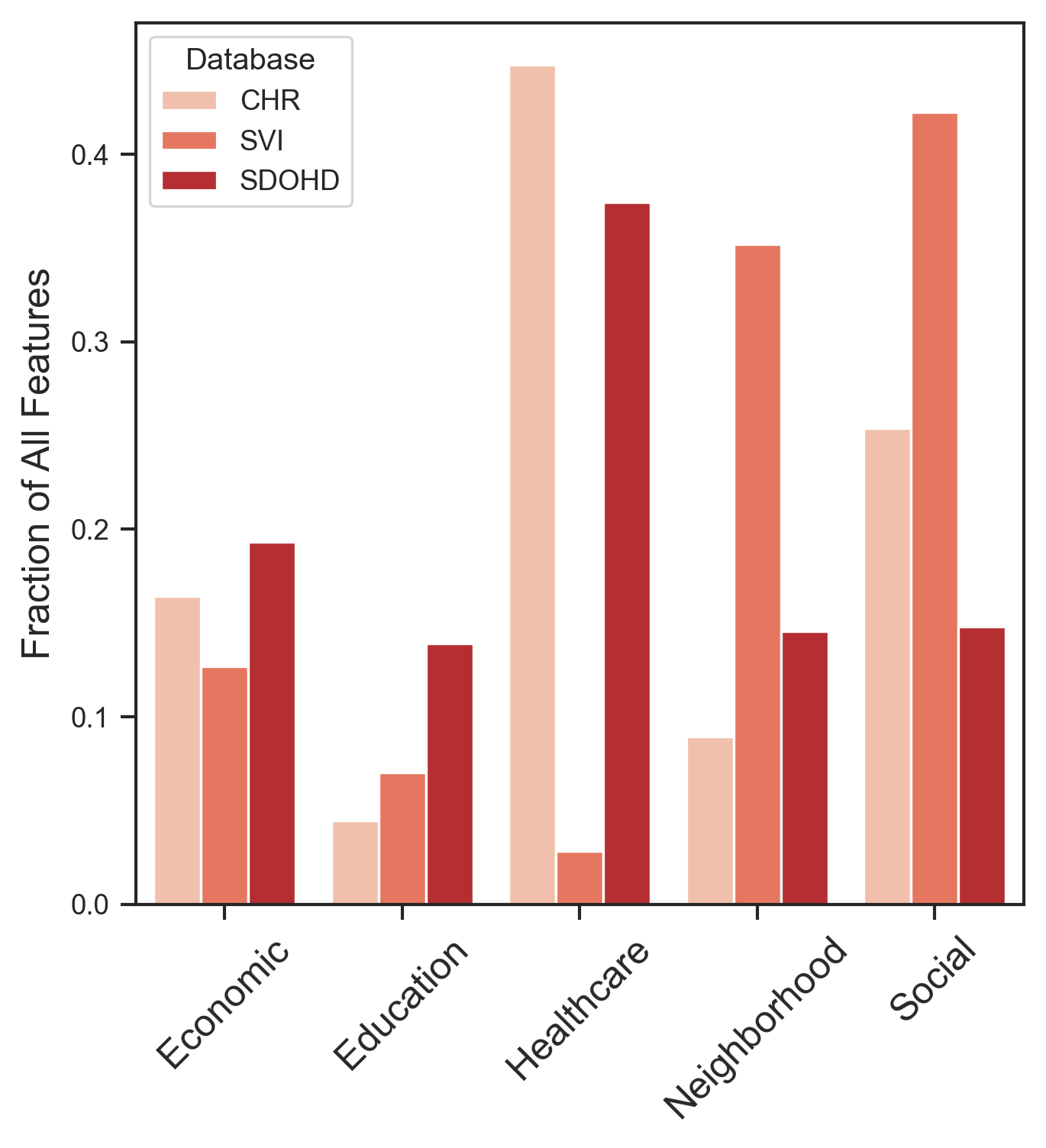}
        \caption{Distribution of features in each SDOH database by domain.}
        \label{fig:dist}
    \end{subfigure}
    
    \caption{Comparison of features in the MIMIC-IV-SDOH databases. SDOHD is the most comprehensive and granular SDOH database of the three. Its features are more correlated with race than CHR and SVI indices. Each database emphasizes a set of SDOH domains more than others.}
    \label{fig:compare_sdoh}
\end{figure*}

\begin{table}[b]
\caption{Characteristics of the final MIMIC-IV-SDOH tables where $d$ is the number of SDOH features. }
\label{tab:sdoh_data}
\resizebox{\columnwidth}{!}{%
\begin{tabular}{@{}cccP{7mm}@{}}
\hline
\begin{tabular}[c]{@{}c@{}}SDOH \\ Database\end{tabular} & Data Version/Year                & \begin{tabular}[c]{@{}c@{}}Geographic \\ Level\end{tabular} & $d$ \\ \hline \hline
CHR   & 2010-2020              & County           & 163              \\
SVI   & 2008, 2014, 2016, 2018 & County     & 160                  \\
SDOHD & 2009-2020              & County     & 1327                \\ \hline
\end{tabular}%
}
\end{table}

\subsection{Public SDOH Databases}
While there exist other SDOH databases, such as Area Deprivation Index \cite{kind_making_2018} and Atlas of Rural and Small-Town America \cite{us_department_of_agriculture_economic_research_service_atlas_2022}, they are either domain-specific or not frequently updated. Because MIMIC-IV contains ICU stays from 2008 to 2019, we focus on databases with SDOH variables that span multiple years and all five SDOH domains, as defined by Healthy People 2030 \cite{healthy_people_2030_social_2020}.

\subsubsection{County Health Rankings (CHR)} 
CHR evaluates counties within each state in the United States based on modifiable health determinants and is updated annually. CHR estimates that clinical care only accounts for 20\% of all contributors to long-term health outcomes, specifically the length and quality of life. The remaining 80\% stems from health behaviors (30\%), physical environment (10\%), and social and economic factors (40\%) \cite{hood_county_2016}. 

\subsubsection{Social Vulnerability Index (SVI)} 
Based on data from the American Community Survey (ACS), SVI evaluates social factors across four themes: socioeconomic status, household composition and disability, minority status and language, and housing type and transportation. Although the index was designed to assess community preparedness and resilience in face of natural hazards, SVI has been used in many population health and health equity studies \cite{karmakar_association_2021,tipirneni_contribution_2022,lehnert_spatial_2020,yu_social_2020,azap_association_2020}. For example, communities with higher levels of social vulnerability experienced greater COVID-19 incidence and mortality \cite{karmakar_association_2021,tipirneni_contribution_2022,howell_associations_2022}. Unlike CHR, SVI is available at both the county level and census tract level.

\subsubsection{Social Determinants of Health Database (SDOHD)} 
To incorporate more granular SDOH data into MIMIC-IV, the last database used in the integration is the Social Determinants of Health Database (SDOHD), which is available at the county, census tract, and zip code levels. The database was recently developed to provide a range of well-documented, readily linkable SDOH variables across domains without having to access multiple source files. SDOHD is curated based on the five key SDOH domains defined by Healthy People 2030: economic stability, education access and quality, health care and quality, neighborhood and built environment, and social and community context.

\subsection{EHR-SDOH Integration}

The creation of the integrated MIMIC-IV-SDOH database is carried out in three steps. 
\paragraph{Step 1: SDOH Data Acquisition} 
For each SDOH database, we concatenate all datasets released between 2008 and 2020. We map each feature to one of the five SDOH domains and provide detailed documentation.  

\paragraph{Step 2: Geographic Crosswalk}
Although SVI and SDOHD are available at the census tract level, we only use county-level data to minimize the risk of patients being identified. Each patient's zip code is mapped to a county using the crosswalk files provided by the United States Department of Housing and Urban Development (HUD) \cite{wilson_understanding_2018}. The files contain a residential ratio column, the ratio of residential addresses in the zip-county area to the total number of residential addresses in the entire zip. Because the mapping is many-to-many, the residential ratio is treated as the probability that the patient with zip code $z$ lives in the census tract $t$ or county $c$, as suggested by the HUD \cite{din_crosswalking_2020}. Note that only patients with Massachusetts zip codes that are in the HUD crosswalk files are included in the final MIMIC-IV-SDOH dataset. 

\paragraph{Step 3: Data Merging} 
MIMIC-IV is merged with each of the three SDOH databases using the geographic location and the SDOH data year closest to the year of admission.

\subsection{Comparison of SDOH Features}
The demographic features in MIMIC-IV, such as race and gender, are sometimes used as proxies for SDOH features, such as socioeconomic status and health behaviors \cite{chen_can_2019,roosli_peeking_2022}. We find that many community-level SDOH features are weakly correlated with race in MIMIC-IV. For example, three SDOHD features, the percentage of households that receive food stamps, the percentage of workers taking public transportation, and the percentage of the population with educational attainment less than high school are all weakly and positively correlated with the Black race (\autoref{fig:heatmap}).

Though to a lesser extent, subindices from SVI (e.g., socioeconomic) and CHR (e.g. health outcomes, quality of life, and social and economic factors) are also weakly associated with race. There are no strong correlations between SDOH features and other tabular features such as labs, risk scores, and Charlson comorbidities.

To better illustrate the type of features in each SDOH database, we manually map each feature to one of the five SDOH domains (\autoref{fig:dist}). The CHR and SDOHD features are predominantly of the Healthcare Access and Quality domain. SVI emphasizes the Neighborhood and the Built Environment domain and the Social and Community Context domain more. 

\section{Data and Methods}
Our primary goal is to determine how incorporating SDOH in ML models could impact predictions of acute and longitudinal outcomes. Leveraging the newly created MIMIC-IV-SDOH database, we assess the impact from the perspective of classification performance and group fairness. We also provide a preliminary investigation of the possible mechanisms behind the contributions of SDOH to model performance or the lack thereof.

\subsection{Data}
In this study, we analyze five patient populations in the MIMIC-IV-SDOH database to assess the impact of SDOH across three tasks. MIMIC-IV data comes from a single EHR system in one geographic location, so the variation in the community-level SDOH features might be too low to be informative. Many past studies that used community-level SDOH features with EHR data from a single hospital or region ended with similar conclusions \cite{chen_social_2020}. To examine the generalizability of our finding, we compare the MIMIC-IV cohort to a patient population in the All of Us Controlled Tier Dataset v6 \cite{the_all_of_us_research_program_investigators_all_2019} for the task of 30-day readmission. Unlike MIMIC-IV, which comes from a single hospital in Boston, the All of Us dataset contains patient-level data from more than 35 hospitals across the United States. Because of this difference, the variation in the SDOH data in All of Us is much greater than that in MIMIC-IV (\autoref{fig:aou_mimic_box}). In addition, based on the distribution of SDOH features, we note that the patients in MIMIC-IV are on average more affluent and more educated than the patients in All of Us.

\subsubsection{MIMIC-IV-SDOH}
Our analysis only includes adult ICU patients (i.e. over 18) from MIMIC-IV v2.2 with a hospital length of stay of at least 3 days. The cohort contains 42,665 patients and a total of 54,380 admissions. 

\paragraph{Task Definition}
We focus on three common classification tasks: (1) in-hospital mortality, (2) 30-day readmission, and (3) one-year mortality. Patients who have expired during a stay are excluded from predictions of 30-day readmission and one-year mortality. For 30-day readmission, we only consider non-elective readmissions. 

\paragraph{Patient Population Definition}
A recent study suggests that the impact of incorporating SDOH in predictive models varies by subpopulation -- vulnerable populations like Black patients and the elderly are likely to benefit more from the inclusion of SDOH \cite{belouali_impact_2022}. Moreover, several studies have suggested that SDOH are strongly associated with glycemic control \cite{walker_relationship_2014}, as well as diabetic risk, morbidity, and mortality \cite{hill-briggs_social_2022}.  
Diabetic patients also use significantly more healthcare resources compared to patients with other chronic diseases \cite{fuchs_icu_2012}. In fact, they account for 31\% of all ICU patients in MIMIC-IV. 

Thus, in addition to the cohort of all ICU patients, we evaluate five subgroups: 
\begin{enumerate}
    \item Diabetic patients
    \item Black diabetic patients
    \item Elderly diabetic patients over 75 years old
    \item Female diabetic patients
    \item Non-English speaking diabetic patients 

\end{enumerate}
On average, all five of these subgroups have more comorbidities compared to the general ICU patients (\autoref{cohort_stats}).

\paragraph{Data Pre-processing}
To better understand the contribution of different types of features to model performance, we divide the entire dataset into a total of 15 feature sets (\autoref{tab:feature_set}) and train separate models on each. These feature sets can be broadly classified into three categories: SDOH features alone, EHR features alone, and SDOH features combined with EHR features. 

\begin{table}[htp]
\centering
\caption{Breakdown of feature sets by the type of EHR data contained}
\label{tab:feature_set}
\begin{tabular}{@{}ll@{}}

\hline 
EHR Data Type      & \multicolumn{1}{l}{Feature Set} \\ \hline \hline
\multirow{3}{*}{No EHR Data/SDOH Alone} & \texttt{CHR}                     \\
                                    & \texttt{SVI}                    \\
                                    & \texttt{SDOHD}                  \\ \hline
\multirow{4}{*}{Structured EHR Data (Tabular)} & \texttt{Tabular}    \\
                                    & \texttt{Tabular+CHR}         \\
                                    & \texttt{Tabular+SVI}         \\
                                    & \texttt{Tabular+SDOHD}        \\ \hline
\multirow{4}{*}{Unstructured EHR Data (Clinical Notes)} & \texttt{Notes}                    \\
                                    & \texttt{Notes+CHR}                      \\
                                    & \texttt{Notes+SVI}                    \\
                                    & \texttt{Notes+SDOHD}             \\ \hline
\multirow{4}{*}{All EHR Data} & \texttt{All} \\
                                    & \texttt{All+CHR}                      \\
                                    & \texttt{All+SVI}                    \\
                                    & \texttt{All+SDOHD}                    \\ \hline
                                    
\end{tabular}
\end{table}

For the EHR features, we include two different data modalities: tabular data and discharge notes. To enable fair comparison across the three tasks, we use the same tabular features and sections of the discharge notes in all prediction tasks. Tabular features include basic demographics, Charlson comorbidities, labs from the first 24 hours of stay, and risk scores (APSIII, SAPS-II, SOFA, and OASIS). The following sections from discharge notes are included: chief complaint, present illness, medical history, medications on admission, allergies, major surgical or invasive procedure, physical exam on admission, pertinent results on admission, and family history. 

Before separating patients into different patient populations, we use median imputation for numerical features before performing standard scaling and constant imputation for categorical features before one-hot encoding. Median imputation is used instead of mean imputation in consideration of skewed data. We also apply principal component analysis (PCA) to reduce the dimensionality of the SDOH data, which is particularly useful as many of the SDOH features are strongly correlated. We retain principal components that explain at least 0.99 of the variance in the data.

Discharge notes are stripped of explicit indicators of in-hospital mortality before being tokenized and lemmatized. Corpus-specific stop words are removed by filtering terms with a document frequency greater than 0.7. Terms with a document frequency less than 0.001 are also removed. Lastly, the notes are converted into a Term Frequency-Inverse Document Frequency (TF-IDF) representation with a vocabulary of size $|V|= 11,751$ words.

\subsubsection{All of Us Cohort}
\label{sec:aou}
Because the All of Us dataset is made up of primarily living participants, we focus on the prediction of 30-day readmission. We only include adult participants who had an in-patient hospital stay between 2014 and 2021. Unlike MIMIC-IV, the hospital stays in All of Us are not limited to the ICU, so these patients have fewer comorbidities on average (\autoref{cohort_stats}).

We exclude patients who stayed less than 3 days in the hospital and didn't have any lab results in the first 24 hours; these are the same labs used in MIMIC-IV. We have 13,324 patients and 21,555 admissions in the final All of Us cohort, representing all 50 US states but Kentucky. More than 50\% of the patients come from the Northeast region. 

The All of Us dataset only has 7 community-level SDOH features sourced from the 2017 ACS via a three-digit zip code linkage (\autoref{apd:aou}). 
Tabular features in the All of Us dataset are the same as those in MIMIC-IV except for clinical risk scores, which are not available. We apply the same data pre-processing techniques on both datasets. Because the All of Us dataset has no clinical notes, we only train models on three feature sets: (\texttt{Tabular}, \texttt{SDOH}, and \texttt{Tabular+SDOH}).

\subsection{Models Benchmarked}
We train three types of machine learning models -- logistic regression \cite{scikit-learn}, random forest classifier \cite{scikit-learn}, and XGBoost classifier \cite{xgboost} -- for each task, patient population, and feature set combination. Each dataset is partitioned into 70:30 train-test splits. To prevent data leakage, no patient appears in both the training set and the test set. Each model is tuned through random hyperparameter search \cite{bergstra_random_2012} under broad parameter distributions. See \autoref{apd:training} for additional training details. 

\begin{figure*}[htp]
    \centering
    \includegraphics[width=\textwidth]{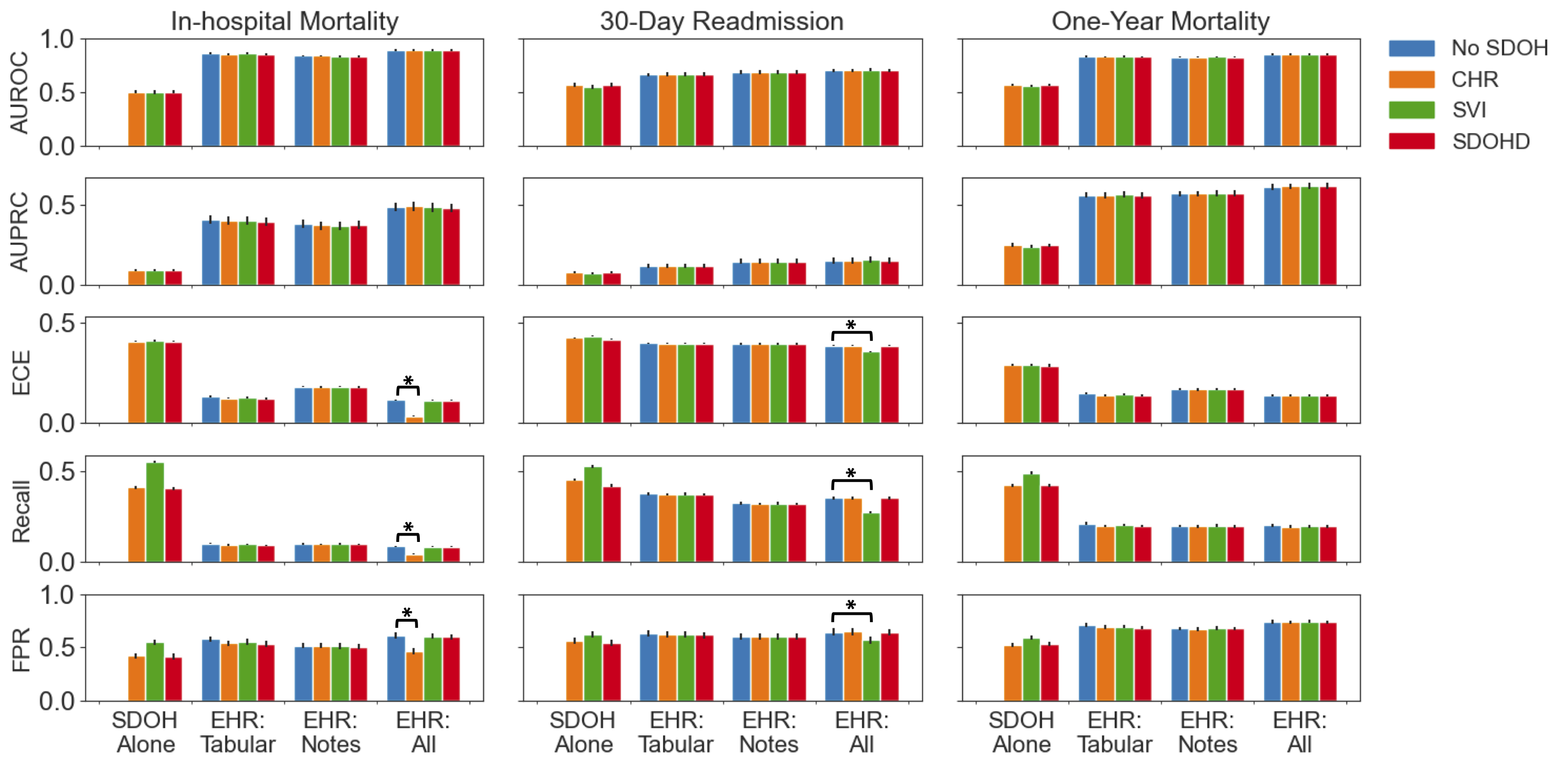}
    \caption{Comparison of the performance of XGBoost classifiers trained on 15 different feature sets to predict in-hospital mortality, 30-day readmission, and one-year mortality in the general ICU population. There are no detectable differences between AUROC and AUPRC of models that do not incorporate SDOH features and those that do. However, when combining SDOH features with all EHR data, we observe significant impacts on ECE, FPR, and recall. We highlight such occurrences with asterisks. The error bars denote the 95\% confidence intervals obtained through 1000 bootstrap samples. }
    \label{fig:all_classification}
\end{figure*}

\subsection{Evaluation}
\subsubsection{Classification Performance}
\label{sec:perf_metrics}
We evaluate the models in terms of three primary metrics: 1) area under the receiver operating characteristic curve (AUROC), 2) area under the precision-recall curve (AUPRC), and 3) expected calibration error (ECE). While AUROC is a standard metric to assess accuracy, we include AUPRC to account for class imbalance and ECE to measure the reliability of the prediction. 
We also use recall as a secondary metric. While threshold selection is complex, cost-dependent, and application-specific, we use a classification threshold of 0.5 for demonstration purposes. 95\% confidence intervals are constructed for all metrics by sampling the test set for 1000 bootstrap iterations \cite{diciccio_bootstrap_1996}.

\subsubsection{Group Fairness} 
\label{sec:fairness_eval}
In addition to classification parity, we evaluate the FPR parity -- based on the equality of opportunity definition of group fairness \cite{hardt_equality_2016}. 
\[ \hat Y \perp\!\!\!\perp G \mid Y = 0 \] 
In other words, the probability of the model predicting a negative outcome is independent of group attribute $G$, conditional on the outcome $Y$ being a true negative.

We examine the differences in TPR across subgroups defined based on the following attributes: (1) race, (2) age, discretized into four bins, (3) gender, (4) median household income, (5) percentage of workers commuting via public transportation, (6) percentage of the population with educational attainment less than high school, (7) percentage of the population receiving food stamps, and (8) the percentage of non-citizens. The SDOH features are discretized into quartiles.

\section{Impact of SDOH on Clinical Prediction Tasks}

\subsection{Impact of SDOH on the General Population}
We first examine the impact of SDOH on the general ICU patient population in MIMIC-IV. We find that the inclusion of community-level SDOH in models does not help with predictive performance, measured by AUROC and AUPRC, but can lead to better-calibrated models with a lower FPR. We validate this finding on a more geographically diverse dataset: the All of Us dataset.

\subsubsection{No Improvement in Model Accuracy}
First, SDOH features alone, without any EHR data, are not predictive of individual patient outcomes. The mean test AUC of the XGBoost models, the best models in terms of AUPRC, trained on SDOH alone is below 0.60 across all tasks, substantially lower than those trained on tabular EHR features and the TF-IDF representation of discharge notes (\autoref{fig:all_classification}). This is not particularly surprising as most studies that utilize community-level SDOH have arrived at similar conclusions \cite{chen_can_2019}. One possible explanation is that community-level estimates are either imprecise or biased, especially if the within-community variance of a feature is high. Moreover, when a patient is critically ill, information on their upstream risk factors might not be as useful as their current state of health.

\begin{table*}[!b]
\caption{Comparison of model performance for XGBoost classifiers trained with and without \texttt{SDOH} for the task of 30-day readmission in MIMIC-IV and the All of Us dataset. In both datasets, the addition of SDOH has no effect on model performance.}
\centering
\label{tab:all_vs_black_diabetic}
\resizebox{0.85\textwidth}{!}{%
\begin{tabular}{l|ccccc|ccccc}
\hline
\multirow{2}{*}{Feature Set} & \multicolumn{5}{c|}{MIMIC-IV} & \multicolumn{5}{c}{All of Us} \\
&    AUROC & AUPRC &   ECE &   FPR & Recall &     AUROC & AUPRC &   ECE &   FPR & Recall       \\
\hline \hline
SDOH         &     0.57 &  0.08 &  0.42 &  0.42 &   0.54 &      0.53 &  0.21 &  0.30 &  0.48 &   0.50 \\
Tabular      &     0.67 &  0.11 &  0.40 &  0.38 &   0.63 &      0.60 &  0.26 &  0.27 &  0.35 &   0.47 \\
Tabular+SDOH &     0.67 &  0.11 &  0.40 &  0.37 &   0.62 &      0.60 &  0.26 &  0.27 &  0.34 &   0.46 \\
\hline
\end{tabular}%
}
\label{tab:aou_vs_mimic}
\end{table*}

\begin{table*}[htp]
\caption{Combinations of EHR and SDOH features that influence performance of the best models for each patient population in terms of AUPRC. We report the number of occurrences in which incorporating SDOH features significantly improves or worsens performance in the form of \texttt{(\# improves/\# worsens)} across the three prediction tasks (total number of occurrences is 3). Significance is evaluated using a 1000-sample bootstrap hypothesis test at the 5\% significance level.}
\label{tab:all_count_improves_worsens}
\resizebox{0.8\textwidth}{!}{%
\begin{tabular}{l|l|ccc|ccc|ccc}
\hline 
\multirow{2}{*}{\begin{tabular}[c]{@{}l@{}}Patient \\ Population\end{tabular}}              & \multirow{2}{*}{Metric} & \multicolumn{3}{c|}{Tabular}                                                          & \multicolumn{3}{c|}{Notes}                                                            & \multicolumn{3}{c}{All}                                                              \\
                                                                                            &                         & CHR                        & SVI                        & SDOHD                      & CHR                        & SVI                        & SDOHD                      & CHR                        & SVI                        & SDOHD                      \\
\hline \hline                                 
\multirow{3}{*}{All Diabetic}                                                               & ECE         & {\color[HTML]{9A0000} 0/1} & 0/0                        & 0/0                        & 0/0                        & 0/0                        & {\color[HTML]{9A0000} 0/1} & 0/0                        & 0/0                        & 0/0                        \\
                                                                                            & FPR         & {\color[HTML]{9A0000} 0/1} & 0/0                        & 0/0                        & 0/0                        & 0/0                        & 0/0                        & 0/0                        & 0/0                        & 0/0                        \\
                                                                                            & Recall       & {\color[HTML]{009901} 1/0} & 0/0                        & 0/0                        & 0/0                        & 0/0                        & 0/0                        & 0/0                        & 0/0                        & 0/0                        \\ \hline 
\multirow{3}{*}{Black Diabetic}                                                             & ECE         & {\color[HTML]{009901} 1/0} & 0/0                        & 0/0                        & 0/0                        & {\color[HTML]{009901} 1/0} & {\color[HTML]{9A0000} 0/1} & 0/0                        & {\color[HTML]{9A0000} 0/1} & {\color[HTML]{9A0000} 0/1} \\
                                                                                            & FPR         & {\color[HTML]{009901} 1/0} & {\color[HTML]{9A0000} 0/1} & 0/0                        & 0/0                        & {\color[HTML]{009901} 1/0} & {\color[HTML]{9A0000} 0/1} & 0/0                        & {\color[HTML]{9A0000} 0/1} & {\color[HTML]{009901} 1/0} \\
                                                                                            & Recall       & {\color[HTML]{9A0000} 0/1} & {\color[HTML]{009901} 1/0} & 0/0                        & 0/0                        & 0/0                        & 0/0                        & 0/0                        & {\color[HTML]{009901} 1/0} & 0/0                        \\ \hline 
\multirow{3}{*}{Elderly Diabetic}                                                           & ECE         & 0/0                        & {\color[HTML]{009901} 2/0} & 0/0                        & 0/0                        & 0/0                        & {\color[HTML]{009901} 1/0} & 0/0                        & 0/0                        & 0/0                        \\
                                                                                            & FPR         & 0/0                        & {\color[HTML]{009901} 1/0} & 0/0                        & 0/0                        & 0/0                        & {\color[HTML]{9A0000} 0/1} & 0/0                        & 0/0                        & 0/0                        \\
                                                                                            & Recall       & 0/0                        & 0/0                        & 0/0                        & 0/0                        & 0/0                        & {\color[HTML]{009901} 1/0} & 0/0                        & 0/0                        & 0/0                        \\ \hline 
\multirow{3}{*}{Female Diabetic}                                                            & ECE         & 0/0                        & 0/0                        & 0/0                        & {\color[HTML]{009901} 1/0} & 0/0                        & 0/0                        & 0/0                        & 0/0                        & 0/0                        \\
                                                                                            & FPR         & 0/0                        & 0/0                        & 0/0                        & {\color[HTML]{009901} 1/0} & 0/0                        & 0/0                        & {\color[HTML]{009901} 1/0} & 0/0                        & 0/0                        \\
                                                                                            & Recall       & 0/0                        & 0/0                        & 0/0                        & {\color[HTML]{9A0000} 0/1} & 0/0                        & 0/0                        & 0/0                        & 0/0                        & 0/0                        \\ \hline 
\multirow{3}{*}{\begin{tabular}[c]{@{}l@{}}Non-English\\ Speaking \\ Diabetic\end{tabular}} & ECE         & 0/0                        & {\color[HTML]{009901} 1/0} & {\color[HTML]{009901} 1/0} & {\color[HTML]{9A0000} 0/1} & 0/0                        & 0/0                        & 0/0                        & 0/0                        & {\color[HTML]{009901} 1/0} \\
                                                                                            & FPR         & 0/0                        & 0/0                        & 0/0                        & {\color[HTML]{9A0000} 0/1} & 0/0                        & 0/0                        & 0/0                        & 0/0                        & 0/0                        \\
                                                                                            & Recall       & 0/0                        & 0/0                        & 0/0                        & 0/0                        & 0/0                        & 0/0                        & 0/0                        & 0/0                        & 0/0                \\ \hline       
\end{tabular}%
}
\end{table*} 

Similarly, combining SDOH with tabular EHR data and discharge notes does not improve the AUROC and AUPRC of the model. Again, this trend is observed in all model classes and SDOH databases. This suggests that the added SDOH features do not provide additional information beyond what is already captured in the EHR. However, SDOH features have some influence on other metrics as such ECE, TPR, and recall. For example, for the task of in-hospital mortality prediction, combining CHR features with all EHR features significantly reduces ECE from 0.11 to 0.03 and FPR from 0.09 to 0.04 (\autoref{apd:best_models}). Likewise, for the task of 30-day readmission, combining SVI features with all EHR features significantly reduces ECE from 0.39 to 0.35 and FPR from 0.35 to 0.27. However, these improvements are at the expense of a lower recall.

\subsubsection{Generalizability of the Finding}
Using the All of Us dataset, we validate that our finding generalizes to a more geographically diverse cohort. Unlike MIMIC-IV, which represents critically ill patient stays at a hospital in Boston, the All of Us cohort includes all in-patient stays across the United States, and many patients do not have any comorbidities (\autoref{cohort_stats}). 
The AUROC of the XGBoost classifiers for All of Us is lower than that in MIMIC due to the lack of detailed, hourly lab and vitals. The AUPRC is higher for All of Us because the 30-day readmission rate is 18\% in the All of Us cohort and only 6\% in the MIMIC-IV cohort.

Because the two cohorts are drastically different, a direct comparison of model performance is not meaningful, but we can examine the trend in the added value of SDOH. 
Consistent with our results in MIMIC-IV, we see no significant performance differences in models trained with tabular EHR data and tabular EHR data combined with SDOH data in the All of Us cohort (\autoref{tab:aou_vs_mimic}).


\subsection{Varying Impact of SDOH on Vulnerable Patient Populations}
In this section, we investigate whether including SDOH in predictive models can lead to better performance for specific patient populations. For each task, we compare the best model that incorporates SDOH data to the best baseline model trained on only EHR data.

\subsubsection{Limited Improvement in Model Performance in Vulnerable Patient Populations}
Similar to the general ICU patients, we find that incorporating SDOH has some impact on model performance in the more vulnerable patient populations. Although predictive performance metrics such as AUROC and AUPRC are largely unaffected, we observe significant improvements in ECE, FPR, or recall in selected models trained on SDOH data. In \autoref{tab:all_count_improves_worsens}, we report the aggregated number of occurrences in which incorporating SDOH features significantly improves or worsens model performance across three prediction tasks. See \autoref{apd:best_models} for more granular results. 

We find that the added value of SDOH features varies by patient population, prediction task, and the EHR features they are combined with. Even for the same patient population and task, SDOH features are not equally informative or useful. For example, for models trained on all diabetic patients, the only observed performance boost is in recall when CHR features are combined with tabular features. However, the improvement in the recall is at the expense of higher ECE and FPR. In contrast, for models trained on female diabetic patients, incorporating CHR improves model performance in at least one of the three metrics when combined with discharge notes or all EHR data but not tabular EHR data alone. 

This varying effect of the inclusion of SDOH by patient population precisely captures why SDOH should be collected and incorporated in analyses. Although individuals in a neighborhood are exposed to the same community-level SDOH, they have varying social needs \cite{agency_for_healthcare_research_and_quality_about_2020}. Incorporating SDOH into predictive models may be helpful to identify patients with specific needs and reduce health disparities associated with poor social conditions \cite{andermann_taking_2016, adler_addressing_2016, chen_can_2019}. 


\section{SDOH As Fairness Audit Categories}
A 2014 report by the National Academies of Medicine (NAM) argued that the integration of SDOH into the EHR would better enable healthcare providers to address health disparities \cite{committee_on_the_recommended_social_and_behavioral_domains_and_measures_forelectronic_health_records_capturing_2015}. Extending on a previous study on the integration of SDOH features and model fairness in patients with heart failure \cite{li_improving_2022}, we conduct a thorough audit of FPR parity in all ICU patients using SDOH features in addition to protected attributes such as race, age, and gender. To enable evaluation based on SDOH features, they are binned into quartiles, and the bin edges are documented in \autoref{apd:sdoh_bins}. 

In \autoref{fig:all_fpr_30_day}, we report the FPR of classifiers with the highest AUPRC for 30-day readmission prediction. We focus on the models trained on all EHR data (\texttt{All}) and all EHR data combined with the most helpful SDOH features (\texttt{All+SVI}) across different subgroups. See \autoref{fig:fpr_all} for the other two prediction tasks.

\subsection{SDOH Features Enable More Granular Audits}
In this setting, a high FPR indicates that the model is overdiagnosing or falsely claiming that the patient is high-risk, which has both medical and economic costs \cite{hoffman_overdiagnosis_2012}. A high FPR disparity means that members of a protected subgroup would not be given the correct diagnosis or appropriate intervention at the same rate as the other patients.  

\begin{figure*}[htp]
    \centering
    
    
        \includegraphics[width=0.9\textwidth]{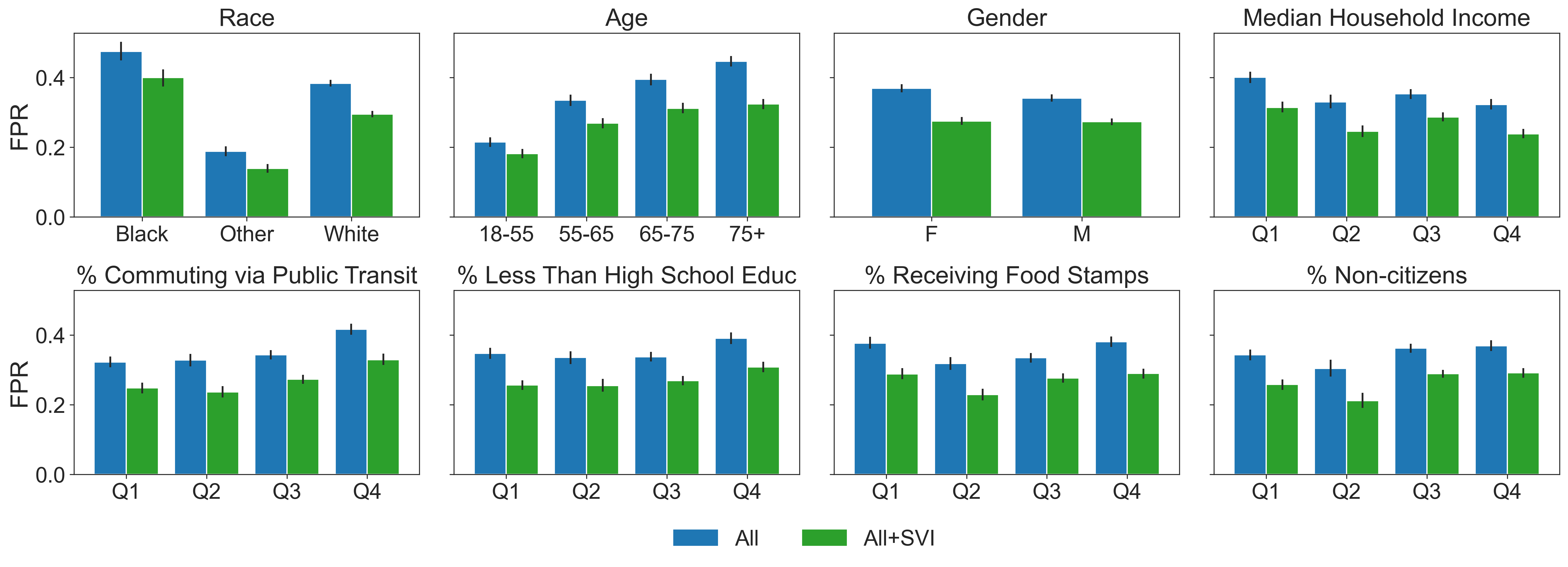}

    
    
    \caption{Comparison of the FPR of XGBoost classifiers trained on all EHR data (\texttt{All}) and all EHR data combined with SVI features for 30-day readmission prediction in all MIMIC-IV ICU patients. FPR is reported for subgroups defined by race, gender, age, and five SDOH features, which are binned into quartiles. The bin edges are documented in \autoref{apd:sdoh_bins}. The error bars denote the 95\% confidence intervals obtained through 1000 bootstrap samples.}
    \label{fig:all_fpr_30_day}
\end{figure*}

An audit of the FPR based on protective attributes confirms findings from prior work that algorithms exhibit biases against underserved patient populations \cite{seyyed-kalantari_underdiagnosis_2021, zhang_hurtful_2020, zhang_improving_2022}. We find that patients who are older, Black, or female have higher FPRs. (\autoref{fig:all_fpr_30_day}). These differences are among the most commonly reported findings in health disparities research; often, these studies stop there without connecting the observed disparities to mechanisms of systemic biases that drive downstream adverse health outcomes \cite{lett_conceptualizing_2022}. This is partially due to the lack of additional information on the patients beyond basic demographics.

The fairness audit based on SDOH features provided additional insight and raised more questions. We find that the FPR is elevated for patients residing in communities where the median household income is low, a larger proportion of individuals commute to work using public transit, and a larger proportion of individuals did not complete high school (\autoref{fig:all_fpr_30_day}). 

We hypothesize that the FPR disparity is a result of bias propagation, which has been suggested by previous studies \cite{seyyed-kalantari_underdiagnosis_2021, adam_write_2022}. While future work is needed to validate the hypothesis, one interpretation of the FPR disparity between patients in quartiles defined based on the percent of workers commuting via public transit is that patients in the fourth quartile likely do not own a car and hence have higher transportation barriers and limited access to healthcare \cite{cristancho_listening_2008,garcia_popa-lisseanu_determinants_2005, syed_traveling_2013, towne_social_2021,cooper_poverty_2012,ney_continuous_2013}. Additionally, a lower household income and lower educational attainment could represent socioeconomic and linguistic disparities in access to care \cite{gottlieb_poverty_1995, towne_social_2021, cristancho_listening_2008, chang_social_2019}.



\section{Discussion}
\subsection{The Need for Better Data}
Overall, our analysis validates previous findings that community-level SDOH features do not improve the accuracy of clinical prediction tasks \cite{chen_social_2020} in both a multi-year cohort and a geographically diverse cohort. We expect individual-level SDOH to be better predictors of outcomes, as prior studies that incorporated them all reported significantly improved performance \cite{takahashi_health_2015,nijhawan_clinical_2019, amrollahi_inclusion_2022}. However, this data are not readily available. For example, although individual-level SDOH can be extracted from participant surveys in the All of Us dataset, less than 15\% of the participants have completed the SDOH survey. Moreover, the survey responses were collected only once for each participant, so these survey-based SDOH features may not accurately reflect the lived experience of the respondents beyond the period the survey was conducted. In light of our findings, we call for further efforts to standardize the routine collection of SDOH data and integration into EHR. 

The healthcare system plays a vital role in collecting, using, and sharing actionable SDOH data \cite{norc_at_the_university_of_chicago_social_2023}. To facilitate this effort, providers and operations staff across care settings should focus on actions that enhance the standardization and integration of SDOH data. Organizations such as the Office of National Coordinator for Health IT (ONC), the Joint Commission, and  Health Level Seven International (HL7) are all leading efforts to further SDOH interoperability and standards \cite{ribick_new_2019, office_of_the_national_coordinator_for_health_information_technology_social_2023}. It should also be a focus to provide sufficient training and education for the staff who are collecting and encoding the data from the patients while adhering to cultural competency, privacy, and confidentiality standards \cite{mcgraw_privacy_2015}.

As the research community awaits access to the more granular EHR-SDOH data, we hope the MIMIC-IV-SDOH database will serve as a starting point for studies on the relationship between community risk factors and patient outcomes and those looking to understand the needs of vulnerable subpopulations.

\subsection{On More Actionable Audits}
Despite spending a higher percentage of our GDP on medical care expenditures than other developed countries, health outcomes in the United States are among the worst for developed countries \cite{oecd_health_2021}. Numerous studies have confirmed the potential of AI in improving health outcomes, but very few tools that were developed have actually helped \cite{will_douglas_heaven_hundreds_2021, avi_goldfarb_why_2022}. A promising direction forward is to look beyond the clinical walls and understand the conditions that affect the health of the people upstream \cite{mcginnis_case_2002}. 

The growing evidence around the association between SDOH and health outcomes calls for targeted action, but there is a lack of consensus on what interventions would work \cite{mcginnis_case_2002}. Progress in evidence-informed policymaking requires a commitment to enhancing our current understanding of how SDOH affect different populations and ways to measure the effectiveness of interventions targeting specific SDOH domains.
Thus, community-level SDOH features are essential for evaluating and monitoring health disparities \cite{purtle_uses_2019}. 
 
By evaluating fairness using intersectional social identities, we could better account for the socially constructed nature of protected attributes such as race and gender. Capturing SDOH provides information on the social processes that created health disparities in the first place \cite{chang_social_2019,hahn_education_2015,lett_conceptualizing_2022}. Audits of biases from the lens of SDOH are also more actionable because these features are not social constructs but modifiable factors that can be addressed \cite{purtle_uses_2019}. Consider transportation, which is one of the SDOH features we used in the fairness audit. Surveys and audits have identified transportation barriers as one of the leading causes of missed or delayed medical appointments, especially in the elderly and those in rural areas \cite{syed_traveling_2013, health_research__educational_trust_social_2017, towne_social_2021}. Health insurance and healthcare delivery organizations are addressing the issue through partnerships with popular ride-share companies to provide non-emergency medical transportation (NEMT) services \cite{powers_shifting_2018, unite_us_lyft_2020}. These programs have decreased costs \cite{powers_nonemergency_2016} and the frequency of urgent care visits \cite{patientengagementhit_lyft_2019}. The development of this intervention would not be possible without an understanding of the underlying SDOH and the population affected. 

 
\subsection{Future Work on SDOH and Health Predictions}
While our work shows that the inclusion of SDOH has minimal impact on three common EHR prediction tasks, they could be more helpful in other tasks and patient groups. Specifically, we did not include any phenotype prediction tasks. Given the associations between SDOH and chronic diseases \cite{amrollahi_inclusion_2022, chin_correlates_1997}, it is possible that SDOH features are good risk predictors for specific comorbidities. In addition, SDOH could be important to account for in the estimation of treatment effects, which several studies have done using the MIMIC database but without SDOH data \cite{hsu_association_2015,zhang_calcium_2015, lan_utilization_2019}.
Likewise, although our study utilized three different model classes, they are all relatively simple. Neural networks could potentially uncover more underlying relationships between SDOH and health outcomes \cite{zheng_development_2020}. 

Regardless of the predictive value of SDOH, it is a good idea to account for them in analyses for more granular benchmarking and evaluation of fairness. For example, MIMIC-IV-SDOH can be mapped to MIMIC-CXR, a large database of chest radiographs with radiology reports. There have been many works that focus on group fairness in the field of medical imaging \cite{zhang_improving_2022, seyyed-kalantari_chexclusion_2020, seyyed-kalantari_underdiagnosis_2021}, which the inclusion of SDOH could contribute to.

\section{Conclusion}
This work advances our understanding of the impact of SDOH on health prediction. First, we develop a new EHR-SDOH dataset by linking a popular EHR database, MIMIC-IV, to public community-level SDOH databases. This dataset can be used to uncover underlying trends between community health and individual health outcomes and provide more benchmarks for evaluating bias and fairness. Second, we demonstrate that incorporating SDOH features in certain vulnerable subgroups can improve model performance. The value of adding SDOH features, however, is dependent on the characteristics of the cohort and the prediction task. Third, we highlight that algorithmic audits conducted through the lens of SDOH are more comprehensive and actionable. However, the lack of access to high-resolution, individual SDOH data is a limitation of the study. To address this, future work should focus on collecting individual-level SDOH features and accounting for them in analyses to address patient needs better and promote health equity.

\begin{acks}
This project is supported by the National Institute of Biomedical Imaging and Bioengineering (NIBIB) of the National Institutes of Health (NIH) under grant number R01-EB017205.

We would also like to acknowledge the All of Us Research Program, which is supported by the National Institutes of Health, Office of the Director: Regional Medical Centers: 1 OT2 OD026549; 1 OT2 OD026554; 1 OT2 OD026557; 1 OT2 OD026556; 1 OT2 OD026550; 1 OT2 OD 026552; 1 OT2 OD026553; 1 OT2 OD026548; 1 OT2 OD026551; 1 OT2 OD026555; IAA \#: AOD 16037; Federally Qualified Health Centers: HHSN 263201600085U; Data and Research Center: 5 U2C OD023196; Biobank: 1 U24 OD023121; The Participant Center: U24 OD023176; Participant Technology Systems Center: 1 U24 OD023163; Communications and Engagement: 3 OT2 OD023205; 3 OT2 OD023206; and Community Partners: 1 OT2 OD025277; 3 OT2 OD025315; 1 OT2 OD025337; 1 OT2 OD025276. In addition, the All of Us Research Program would not be possible without the partnership of its participants.

Finally, we would like to thank Hammaad Adams, Aparna Balagopalan, Hyewon Jeong, Qixuan (Alice) Jin, Intae Moon, Vinith Suriyakumar, Yuxin Xiao, Haoran Zhang, Dana Moukheiber, Lama Moukheiber, and Mira Moukheiber for their valuable feedback and constructive review of this work.
\end{acks}

\bibliographystyle{ACM-Reference-Format}
\bibliography{zotero, software}


\appendix

\section{Additional Information on Model Training and Data Pre-processing}
\subsection{Model Training}
\label{apd:training}
Due to the class imbalance in all the prediction tasks, we use AUROC for model selection during hyperparameter tuning. The distributions of parameters sampled during the randomized search for logistic regression (\texttt{lr}), random forest (\texttt{rf}), and XGBoost (\texttt{xgb}) classifiers are as followed:
\lstset{language=Python}
\begin{lstlisting}
lr_param_grid = {
  "C" : [1e-5, 1e-4, 1e-3, 1e-2, 0.1, 1, 5],
  "solver" : ["liblinear"],
}

rf_param_grid = {
  "n_estimators": [50,100,200,500],
  "max_depth": scipy.stats.randint(2, 10),
  "min_samples_split": scipy.stats.randint(2, 10),
  "min_samples_leaf": scipy.stats.randint(1, 10),
}

xgb_param_grid = {
  "n_estimators": [50,100,200,500],
  "max_depth": scipy.stats.randint(2, 10),
  "learning_rate": (0.01,0.05,0.1,0.2,0.3),
  "min_child_weight": scipy.stats.randint(2, 10),
  "colsample_bytree": [0.5,1],
  "subsample" : [0.3,0.6,0.9], 
  "reg_alpha" : scipy.stats.randint(0, 10), 
  "reg_lambda": scipy.stats.randint(0, 10),
}
\end{lstlisting}

\subsection{Binning SDOHD Features}
\label{apd:sdoh_bins}
The quartile bin edges for SDOH features used in the fairness audit are as followed:
 \begin{enumerate}
    \item Percentage of non-citizens: \\ 0, 0.32, 1.09, 2.54, 30.58
    \item Median household income in dollars: \\ 10446, 60698.5, 74902, 92381, 250001
    \item Percentage with less than high school education: \\ 0, 4.04, 6.79, 11.64, 67.49
    \item Percentage of households receiving food stamps: \\ 0, 4.15, 6.85, 12.1, 78.43
    \item Percentage of workers taking public transit:\\ 0, 4.74, 10.5, 20.44, 77.61
 \end{enumerate}
 

\onecolumn
\subsection{SDOH Features in the All of Us Dataset}
\label{apd:aou}

The community-SDOH features used include the following:
\begin{enumerate}
    \item Percentage of households receiving food stamps
    \item Percentage of the population with at least a high school \\diploma
    \item Median household income 
    \item Percentage of the population with no health insurance
    \item Percentage of the population in poverty
    \item Percentage of houses that are vacant
    \item Deprivation index
\end{enumerate}

\setcounter{figure}{0}
\renewcommand{\thefigure}{B\arabic{figure}}
\setcounter{table}{0}
\renewcommand{\thetable}{B\arabic{table}}
\section{Additional Figures and Tables}
\begin{figure*}[htp]
    \centering
\includegraphics[width=0.8\textwidth]{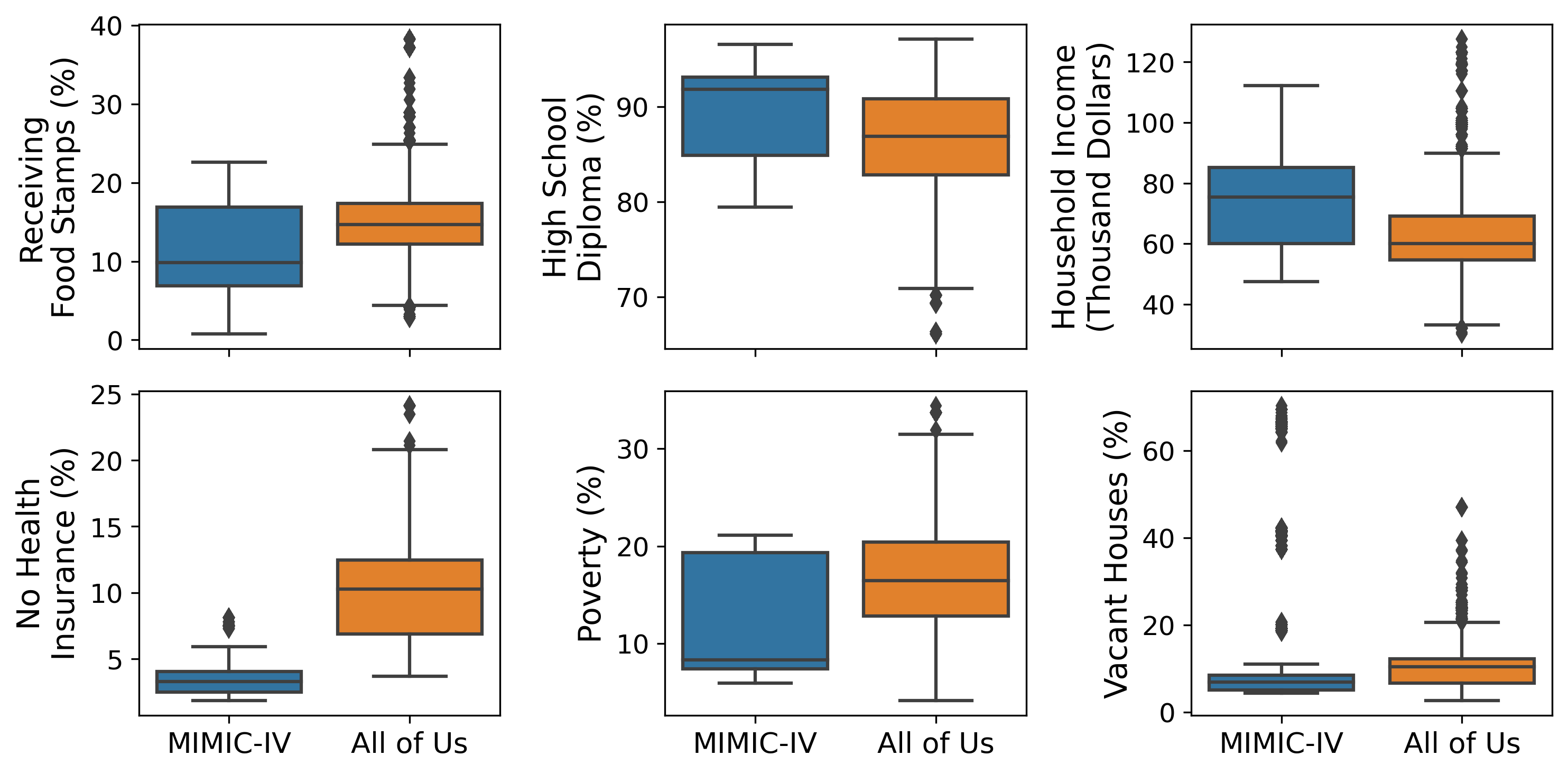}
    \caption{Comparison of selected SDOH features between the MIMIC-IV and the All of Us patient cohorts. Because the All of Us dataset is more geographically diverse, the variation in its SDOH data is much greater than that in MIMIC-IV.}
    \label{fig:aou_mimic_box}
\end{figure*}

\begin{figure*}[htp]
    \centering
    \begin{subfigure}[b]{0.9\textwidth}
        \includegraphics[width=\textwidth]{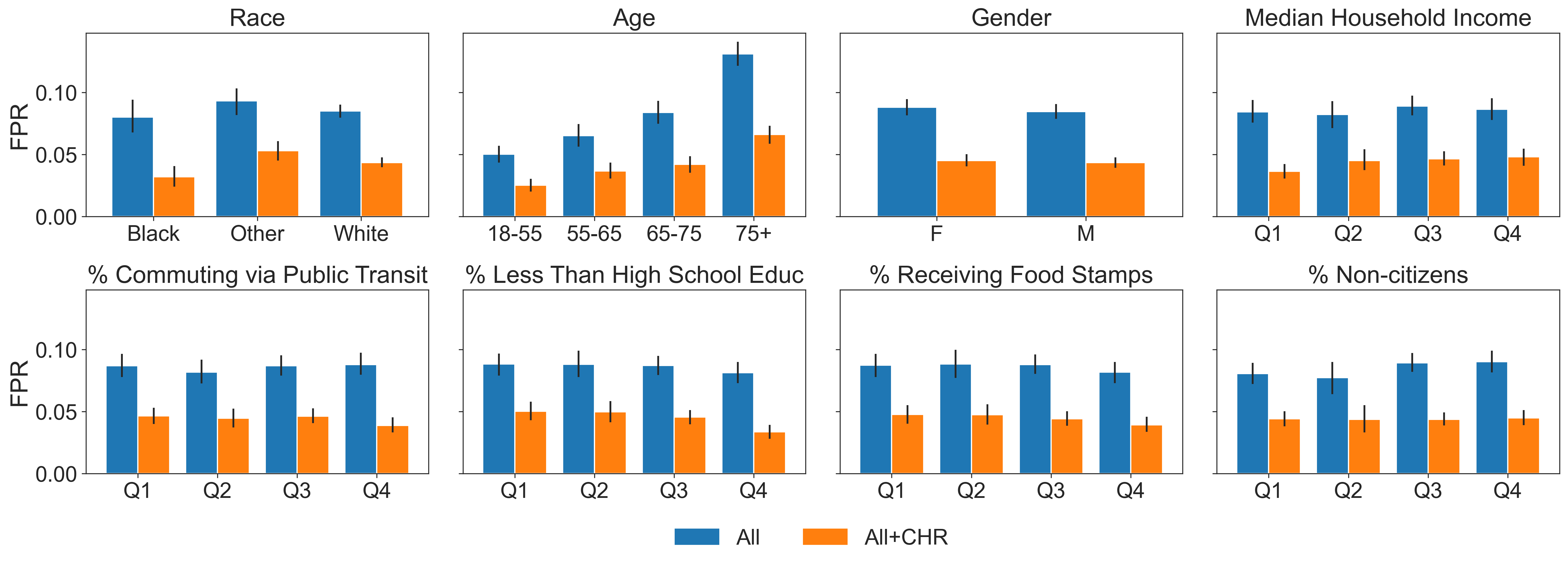}
    \caption{All ICU Patients: In-hospital Mortality}
    \end{subfigure}
    
    \vspace*{0.5cm}
    
    \begin{subfigure}[b]{0.9\textwidth}
        \includegraphics[width=\textwidth]{figures/all_fpr_0.5_by_subgroups_30_day_read.png}
    \caption{All ICU Patients: 30-Day Readmission}
    \end{subfigure}

        \vspace*{0.5cm}
    
    \begin{subfigure}[b]{0.9\textwidth}
        \includegraphics[width=\textwidth]{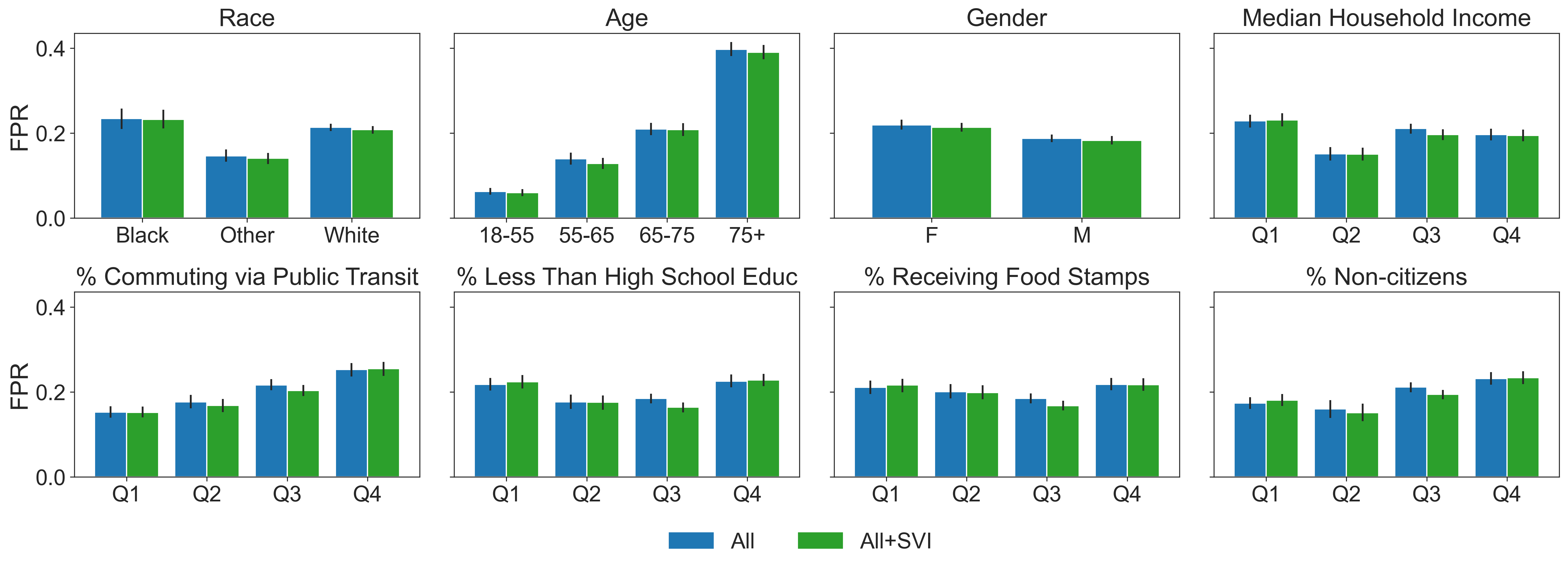}
    \caption{All ICU Patients: One-Year Mortality}
    \end{subfigure}
    
    \caption{Comparison of the FPR of XGBoost classifiers trained on all EHR data (\texttt{All}) and all EHR data combined with best SDOH features for the three tasks in all MIMIC-IV ICU patients. FPR is reported for subgroups defined by race, gender, age, and five SDOH features, which are binned into quartiles. The bin edges are documented in \autoref{apd:sdoh_bins}. The error bars denote the 95\% confidence intervals obtained through 1000 bootstrap samples.}
    \label{fig:fpr_all}
\end{figure*}

\onecolumn
\begin{table}[htp]
\renewcommand{\arraystretch}{1.15}
\caption{Charactertics of the MIMIC-IV and All of Us cohorts. The MIMIC-IV cohort has six patient populations: all ICU patients, diabetic patients, black diabetic patients, elderly diabetic patients, female diabetic patients, and non-English speaking diabetic patients. The All of Us cohort contains all in-patient hospital stays. $N$ is the number of patients in each group.}
\label{cohort_stats}
\resizebox{\columnwidth}{!}{%
\begin{tabular}{@{}|l|l|rrrrrr|r|@{}}
\hline
\multicolumn{1}{|l}{\multirow{4}{*}{Attribute}} & \multicolumn{1}{|l|}{\multirow{4}{*}{Subgroup}} &  \multicolumn{6}{c}{\underline{MIMIC-IV}}  & \multicolumn{1}{|c|}{\underline{All of Us}} \\
& & \multicolumn{1}{c}{\begin{tabular}[c]{@{}c@{}}All ICU \\ (N=42,665)\end{tabular}} & \multicolumn{1}{c}{\begin{tabular}[c]{@{}c@{}}All Diab. \\ (N=12,651)\end{tabular}} & \multicolumn{1}{c}{\begin{tabular}[c]{@{}c@{}}Black Diab. \\ (N=1,710)\end{tabular}} & \multicolumn{1}{c}{\begin{tabular}[c]{@{}c@{}}Elderly Diab. \\ (N=4,520)\end{tabular}} & \multicolumn{1}{c}{\begin{tabular}[c]{@{}c@{}}Female Diab. \\ (N=5,251)\end{tabular}} & \multicolumn{1}{c}{\begin{tabular}[c]{@{}c@{}}Non-English \\ Speaking Diab. \\ (N=1,675)\end{tabular}} & \multicolumn{1}{|c|}{\begin{tabular}[c]{@{}c@{}}All Inpatient \\ (N=13,324)\end{tabular}} \\ \hline \hline
\multirow{4}{*}{Age}                                                                      & 17-55    & 10,136 (23\%)   & 1,806 (14\%)         & 337 (19\%)            & 0 (0\%)                 & 681 (12\%)             & 158 (9\%)                            & 6,481 (48\%)      \\
                                                                                          & 55-65    & 8,773 (20\%)    & 2,759 (21\%)         & 407 (23\%)            & 0 (0\%)                 & 1,008 (19\%)           & 332 (19\%)                           & 3,134 (23\%)      \\
                                                                                          & 65-75    & 10,013 (23\%)   & 3,706 (29\%)         & 473 (27\%)            & 0 (0\%)                 & 1,457 (27\%)           & 415 (24\%)                           & 2,430 (18\%)      \\
                                                                                          & 75+      & 13,742 (32\%)   & 4,380 (34\%)         & 493 (28\%)            & 4,520 (100\%)           & 2,105 (40\%)           & 770 (45\%)                           & 1,279 (9\%)       \\ \hline
\multirow{3}{*}{Gender}                                                                   & Female   & 18,677 (43\%)   & 5,251 (41\%)         & 927 (54\%)            & 2,169 (47\%)            & 5,251 (100\%)          & 801 (47\%)                           & 8,226 (61\%)      \\
                                                                                          & Male     & 23,988 (56\%)   & 7,400 (58\%)         & 783 (45\%)            & 2,351 (52\%)            & 0 (0\%)                & 874 (52\%)                           & 4,800 (36\%)      \\
                                                                                          & Other    & --              & --                   & --                    & --                      & --                     & --                                   & 298 (2\%)         \\ \hline
\multirow{3}{*}{Race}                                                                     & White    & 29,148 (68\%)   & 8,033 (63\%)         & 0 (0\%)               & 3,033 (67\%)            & 3,150 (59\%)           & 455 (27\%)                           & 6,168 (46\%)      \\
                                                                                          & Black    & 3,880 (9\%)     & 1,700 (13\%)         & 1,710 (100\%)         & 523 (11\%)              & 920 (17\%)             & 192 (11\%)                           & 3,589 (26\%)      \\
                                                                                          & Other    & 9,637 (22\%)    & 2,918 (23\%)         & 0 (0\%)               & 964 (21\%)              & 1,181 (22\%)           & 1,028 (61\%)                         & 3,567 (26\%)      \\ \hline
\multirow{3}{*}{Insurance Type}                                                           & Medicaid & 2,976 (6\%)     & 763 (6\%)            & 153 (8\%)             & 68 (1\%)                & 370 (7\%)              & 249 (14\%)                           & 7,109 (53\%)      \\
                                                                                          & Medicare & 18,844 (44\%)   & 6,453 (51\%)         & 781 (45\%)            & 3,169 (70\%)            & 2,828 (53\%)           & 721 (43\%)                           & 3,270 (24\%)      \\
                                                                                          & Other    & 20,845 (48\%)   & 5,435 (42\%)         & 776 (45\%)            & 1,283 (28\%)            & 2,053 (39\%)           & 705 (42\%)                           & 2,945 (22\%)      \\ \hline
\multirow{2}{*}{Language}                                                                 & English  & 38,291 (89\%)   & 11,018 (87\%)        & 1,523 (89\%)          & 3,744 (82\%)            & 4,463 (84\%)           & 0 (0\%)                              & --                \\
                                                                                          & Other    & 4,374 (10\%)    & 1,633 (12\%)         & 187 (10\%)            & 776 (17\%)              & 788 (15\%)             & 1,675 (100\%)                        & --                \\ \hline
\multirow{8}{*}{\begin{tabular}[c]{@{}l@{}}Charlson \\ Comorbidity \\ Index\end{tabular}} & 0        & 2,856 (6\%)     & 0 (0\%)              & 0 (0\%)               & 0 (0\%)                 & 0 (0\%)                & 0 (0\%)                              & 5,977 (44\%)      \\
                                                                                          & 1        & 3,208 (7\%)     & 301 (2\%)            & 63 (3\%)              & 0 (0\%)                 & 110 (2\%)              & 27 (1\%)                             & 1,315 (9\%)       \\
                                                                                          & 2        & 4,232 (9\%)     & 626 (4\%)            & 106 (6\%)             & 0 (0\%)                 & 254 (4\%)              & 56 (3\%)                             & 1,240 (9\%)       \\
                                                                                          & 3        & 5,151 (12\%)    & 1,039 (8\%)          & 131 (7\%)             & 0 (0\%)                 & 397 (7\%)              & 109 (6\%)                            & 1,174 (8\%)       \\
                                                                                          & 4        & 5,576 (13\%)    & 1,434 (11\%)         & 176 (10\%)            & 173 (3\%)               & 561 (10\%)             & 174 (10\%)                           & 855 (6\%)         \\
                                                                                          & 5        & 5,329 (12\%)    & 1,793 (14\%)         & 212 (12\%)            & 511 (11\%)              & 777 (14\%)             & 245 (14\%)                           & 651 (4\%)         \\
                                                                                          & 6        & 4,558 (10\%)    & 1,713 (13\%)         & 215 (12\%)            & 653 (14\%)              & 720 (13\%)             & 230 (13\%)                           & 554 (4\%)         \\
                                                                                          & 7+       & 11,755 (27\%)   & 5,745 (45\%)         & 807 (47\%)            & 3,183 (70\%)            & 2,432 (46\%)           & 834 (49\%)                           & 1,558 (11\%)      \\ \hline 
\end{tabular} %
}
\end{table}

\newpage
\begin{landscape}
\begin{table}[htp]
\renewcommand{\arraystretch}{1.15}
\centering
\caption{Comparison of the performance of models trained with and without SDOH feature to predict in-hospital mortality, 30-day readmission, and one-year mortality for the six patient populations. The evaluation is done only on the best model in terms of AUPRC for each feature set category and task. In general, incorporating SDOH features has a limited impact on model performance. Values in bold indicate significantly better performance compared to the baseline model trained without SDOH, evaluated using a 1000-sample bootstrap hypothesis test at the 5\% significance level.}
\label{apd:best_models}
\resizebox{\columnwidth}{!}{%
\begin{tabular}{|l|l|ccccc|ccccc|ccccc|}
\hline
\multirow{2}{*}{Patient Group} &
  \multirow{2}{*}{Feature Set} &
  \multicolumn{5}{c}{\underline{In-hospital Mortality}} &
  \multicolumn{5}{|c|}{\underline{30-Day Readmission}} &
  \multicolumn{5}{c|}{\underline{One-Year Mortality}} \\
 &
   &
  AUROC ($\uparrow$) &
  AUPRC ($\uparrow$) &
  ECE ($\downarrow$) &
  FPR ($\downarrow$) &
  Recall ($\uparrow$) &
  AUROC ($\uparrow$) &
  AUPRC ($\uparrow$) &
  ECE ($\downarrow$) &
  FPR ($\downarrow$) &
  Recall ($\uparrow$) &
  AUROC ($\uparrow$) &
  AUPRC ($\uparrow$) &
  ECE ($\downarrow$) &
  FPR ($\downarrow$) &
  Recall ($\uparrow$) \\ \hline \hline
\multirow{6}{*}{All ICU} & Tabular &              0.86$^3$ &   0.4$^3$ &           0.13$^3$ &            0.1$^3$ &  0.58$^3$ &           0.66$^3$ &  0.12$^3$ &            0.4$^3$ &           0.37$^3$ &           0.58$^3$ &           0.83$^3$ &  0.56$^3$ &           0.15$^3$ &           0.21$^3$ &           0.71$^3$ \\
                              & Tabular+SDOH &              0.86$^2$ &   0.4$^2$ &           0.13$^2$ &            0.1$^2$ &  0.56$^2$ &           0.67$^3$ &  0.12$^3$ &           0.39$^3$ &           0.36$^3$ &           0.62$^3$ &           0.83$^2$ &  0.56$^2$ &           0.14$^2$ &            0.2$^2$ &           0.69$^2$ \\
                              & Notes &              0.84$^3$ &  0.38$^3$ &           0.18$^3$ &            0.1$^3$ &  0.52$^3$ &           0.69$^3$ &  0.14$^3$ &           0.39$^3$ &           0.32$^3$ &            0.6$^3$ &           0.83$^3$ &  0.57$^3$ &           0.17$^3$ &            0.2$^3$ &           0.68$^3$ \\
                              & Notes+SDOH &              0.84$^3$ &  0.37$^3$ &           0.18$^3$ &            0.1$^3$ &  0.51$^3$ &           0.69$^3$ &  0.14$^3$ &           0.39$^3$ &           0.32$^3$ &            0.6$^3$ &           0.83$^2$ &  0.57$^2$ &           0.17$^2$ &            0.2$^2$ &           0.68$^2$ \\
                              & All &               0.9$^3$ &  0.49$^3$ &           0.11$^3$ &           0.09$^3$ &  0.61$^3$ &            0.7$^3$ &  0.15$^3$ &           0.39$^3$ &           0.35$^3$ &           0.65$^3$ &           0.85$^3$ &  0.61$^3$ &           0.14$^3$ &            0.2$^3$ &           0.74$^3$ \\
                              & All+SDOH &              0.89$^1$ &  0.49$^1$ &  \textbf{0.03}$^1$ &  \textbf{0.04}$^1$ &  0.47$^1$ &           0.71$^2$ &  0.15$^2$ &  \textbf{0.35}$^2$ &  \textbf{0.27}$^2$ &           0.57$^2$ &           0.86$^2$ &  0.62$^2$ &           0.14$^2$ &            0.2$^2$ &           0.74$^2$ \\
\cline{1-17}
\multirow{6}{*}{All Diabetic} & Tabular &              0.86$^3$ &   0.4$^3$ &           0.09$^3$ &           0.07$^3$ &  0.45$^3$ &           0.65$^3$ &  0.15$^3$ &           0.38$^3$ &           0.37$^3$ &           0.55$^3$ &           0.78$^3$ &  0.52$^3$ &           0.13$^3$ &           0.23$^3$ &           0.62$^3$ \\
                              & Tabular+SDOH &              0.85$^2$ &  0.39$^2$ &           0.08$^2$ &           0.06$^2$ &  0.42$^2$ &           0.64$^2$ &  0.14$^2$ &           0.38$^2$ &           0.36$^2$ &           0.58$^2$ &           0.78$^1$ &  0.52$^1$ &           0.19$^1$ &           0.32$^1$ &  \textbf{0.74}$^1$ \\
                              & Notes &              0.82$^3$ &  0.33$^3$ &           0.21$^3$ &           0.18$^3$ &  0.65$^3$ &           0.67$^3$ &  0.17$^3$ &           0.37$^3$ &            0.3$^3$ &           0.55$^3$ &           0.78$^3$ &  0.51$^3$ &           0.14$^3$ &           0.19$^3$ &           0.58$^3$ \\
                              & Notes+SDOH &              0.81$^1$ &  0.32$^1$ &           0.21$^1$ &           0.18$^1$ &  0.63$^1$ &           0.66$^3$ &  0.17$^3$ &           0.39$^3$ &            0.3$^3$ &           0.55$^3$ &           0.78$^1$ &  0.52$^1$ &           0.13$^1$ &           0.18$^1$ &           0.56$^1$ \\
                              & All &              0.89$^3$ &  0.47$^3$ &           0.07$^3$ &           0.05$^3$ &  0.48$^3$ &           0.68$^3$ &  0.18$^3$ &           0.37$^3$ &            0.3$^3$ &           0.54$^3$ &           0.81$^3$ &  0.57$^3$ &           0.11$^3$ &            0.2$^3$ &           0.65$^3$ \\
                              & All+SDOH &              0.89$^2$ &  0.47$^2$ &           0.07$^2$ &           0.05$^2$ &  0.47$^2$ &           0.67$^2$ &  0.18$^2$ &           0.37$^2$ &           0.31$^2$ &           0.52$^2$ &           0.81$^2$ &  0.57$^2$ &           0.11$^2$ &           0.19$^2$ &           0.64$^2$ \\
\cline{1-17}
\multirow{6}{*}{Black Diabetic} & Tabular &              0.83$^3$ &  0.36$^3$ &           0.27$^3$ &           0.22$^3$ &  0.68$^3$ &           0.59$^3$ &  0.14$^3$ &           0.37$^3$ &           0.19$^3$ &           0.31$^3$ &           0.74$^3$ &  0.49$^3$ &           0.18$^3$ &           0.27$^3$ &           0.59$^3$ \\
                              & Tabular+SDOH &              0.83$^2$ &  0.35$^2$ &           0.26$^2$ &            0.2$^2$ &  0.67$^2$ &           0.59$^2$ &  0.16$^2$ &           0.35$^2$ &           0.38$^2$ &  \textbf{0.56}$^2$ &           0.72$^1$ &  0.46$^1$ &  \textbf{0.08}$^1$ &  \textbf{0.17}$^1$ &           0.41$^1$ \\
                              & Notes &              0.76$^3$ &  0.15$^3$ &           0.16$^3$ &           0.06$^3$ &  0.19$^3$ &            0.6$^3$ &  0.17$^3$ &            0.3$^3$ &           0.01$^3$ &           0.02$^3$ &           0.75$^3$ &  0.53$^3$ &           0.21$^3$ &           0.09$^3$ &           0.35$^3$ \\
                              & Notes+SDOH &              0.76$^2$ &  0.19$^2$ &  \textbf{0.05}$^2$ &  \textbf{0.01}$^2$ &  0.07$^2$ &           0.57$^3$ &  0.15$^3$ &           0.26$^3$ &           0.17$^3$ &   \textbf{0.2}$^3$ &           0.74$^1$ &  0.52$^1$ &           0.21$^1$ &           0.08$^1$ &           0.32$^1$ \\
                              & All &              0.83$^3$ &  0.34$^3$ &           0.27$^3$ &           0.21$^3$ &  0.68$^3$ &           0.58$^3$ &  0.15$^3$ &           0.26$^3$ &           0.17$^3$ &           0.24$^3$ &           0.79$^3$ &  0.59$^3$ &           0.21$^3$ &           0.13$^3$ &           0.52$^3$ \\
                              & All+SDOH &              0.82$^2$ &  0.33$^2$ &           0.26$^2$ &           0.21$^2$ &  0.66$^2$ &            0.6$^3$ &  0.18$^3$ &           0.37$^3$ &  \textbf{0.06}$^3$ &           0.14$^3$ &           0.78$^2$ &  0.58$^2$ &           0.27$^2$ &           0.26$^2$ &  \textbf{0.66}$^2$ \\
\cline{1-17}
\multirow{6}{*}{Elderly Diabetic} & Tabular &              0.79$^3$ &  0.38$^3$ &           0.16$^3$ &           0.06$^3$ &  0.35$^3$ &           0.63$^3$ &  0.14$^3$ &           0.37$^3$ &           0.36$^3$ &           0.58$^3$ &           0.73$^3$ &  0.55$^3$ &           0.14$^3$ &           0.31$^3$ &           0.64$^3$ \\
                              & Tabular+SDOH &               0.8$^2$ &  0.37$^2$ &  \textbf{0.08}$^2$ &           0.06$^2$ &  0.36$^2$ &           0.63$^1$ &  0.16$^1$ &           0.37$^1$ &           0.35$^1$ &           0.53$^1$ &           0.73$^2$ &  0.54$^2$ &  \textbf{0.08}$^2$ &  \textbf{0.25}$^2$ &           0.55$^2$ \\
                              & Notes &              0.76$^3$ &  0.29$^3$ &           0.09$^3$ &           0.03$^3$ &  0.17$^3$ &           0.57$^3$ &   0.1$^3$ &           0.37$^3$ &           0.03$^3$ &           0.05$^3$ &           0.72$^3$ &  0.54$^3$ &           0.09$^3$ &           0.23$^3$ &           0.51$^3$ \\
                              & Notes+SDOH &              0.76$^3$ &  0.29$^3$ &           0.09$^3$ &           0.03$^3$ &  0.18$^3$ &           0.55$^3$ &   0.1$^3$ &  \textbf{0.33}$^3$ &            0.2$^3$ &  \textbf{0.24}$^3$ &           0.72$^2$ &  0.55$^2$ &           0.09$^2$ &           0.23$^2$ &           0.53$^2$ \\
                              & All &              0.84$^3$ &  0.43$^3$ &           0.05$^3$ &           0.04$^3$ &  0.31$^3$ &           0.61$^3$ &  0.12$^3$ &           0.25$^3$ &           0.18$^3$ &           0.33$^3$ &           0.76$^3$ &  0.59$^3$ &           0.07$^3$ &           0.22$^3$ &           0.56$^3$ \\
                              & All+SDOH &              0.84$^1$ &  0.43$^1$ &           0.05$^1$ &           0.04$^1$ &  0.31$^1$ &            0.6$^1$ &  0.12$^1$ &           0.25$^1$ &            0.2$^1$ &           0.36$^1$ &           0.76$^2$ &  0.59$^2$ &           0.07$^2$ &           0.21$^2$ &           0.58$^2$ \\
\cline{1-17}
\multirow{6}{*}{Female Diabetic} & Tabular &              0.84$^3$ &   0.4$^3$ &           0.26$^3$ &           0.22$^3$ &  0.73$^3$ &           0.61$^3$ &  0.13$^3$ &           0.37$^3$ &           0.36$^3$ &           0.55$^3$ &           0.75$^3$ &  0.44$^3$ &            0.2$^3$ &           0.32$^3$ &            0.7$^3$ \\
                              & Tabular+SDOH &              0.83$^2$ &  0.38$^2$ &           0.26$^2$ &           0.22$^2$ &  0.71$^2$ &           0.57$^1$ &  0.12$^1$ &           0.35$^1$ &           0.34$^1$ &           0.47$^1$ &           0.75$^3$ &  0.45$^3$ &           0.22$^3$ &           0.35$^3$ &           0.71$^3$ \\
                              & Notes &              0.76$^3$ &  0.28$^3$ &           0.07$^3$ &           0.02$^3$ &  0.12$^3$ &           0.62$^3$ &  0.15$^3$ &            0.4$^3$ &           0.15$^3$ &           0.27$^3$ &           0.75$^3$ &  0.44$^3$ &           0.21$^3$ &            0.3$^3$ &           0.68$^3$ \\
                              & Notes+SDOH &              0.76$^2$ &  0.27$^2$ &           0.07$^2$ &           0.01$^2$ &  0.11$^2$ &           0.61$^1$ &  0.15$^1$ &           0.39$^1$ &           0.13$^1$ &           0.23$^1$ &           0.74$^1$ &  0.44$^1$ &  \textbf{0.12}$^1$ &  \textbf{0.15}$^1$ &           0.41$^1$ \\
                              & All &              0.86$^3$ &  0.44$^3$ &           0.04$^3$ &           0.03$^3$ &  0.38$^3$ &           0.62$^3$ &  0.16$^3$ &           0.39$^3$ &           0.16$^3$ &           0.26$^3$ &           0.77$^3$ &  0.49$^3$ &           0.11$^3$ &           0.18$^3$ &           0.53$^3$ \\
                              & All+SDOH &              0.86$^2$ &  0.44$^2$ &           0.04$^2$ &           0.03$^2$ &  0.37$^2$ &           0.62$^1$ &  0.14$^1$ &           0.38$^1$ &  \textbf{0.08}$^1$ &           0.17$^1$ &           0.77$^3$ &  0.49$^3$ &            0.1$^3$ &           0.17$^3$ &           0.52$^3$ \\
\cline{1-17}    \multirow{6}{*}{\begin{tabular}[c]{@{}l@{}}Non-English \\ Speaking Diabetic\end{tabular}} & Tabular &               0.8$^3$ &  0.34$^3$ &           0.27$^3$ &           0.24$^3$ &  0.68$^3$ &           0.61$^3$ &  0.12$^3$ &           0.36$^3$ &           0.27$^3$ &           0.46$^3$ &            0.8$^3$ &  0.55$^3$ &           0.13$^3$ &           0.13$^3$ &           0.51$^3$ \\
                              & Tabular+SDOH &              0.79$^2$ &  0.35$^2$ &           0.26$^2$ &           0.23$^2$ &  0.71$^2$ &           0.58$^3$ &  0.13$^3$ &   \textbf{0.3}$^3$ &           0.31$^3$ &           0.48$^3$ &           0.79$^2$ &  0.54$^2$ &  \textbf{0.07}$^2$ &           0.14$^2$ &           0.51$^2$ \\
                              & Notes &              0.74$^3$ &  0.27$^3$ &           0.05$^3$ &           0.02$^3$ &  0.08$^3$ &           0.63$^3$ &  0.11$^3$ &           0.29$^3$ &           0.18$^3$ &           0.27$^3$ &           0.71$^3$ &  0.48$^3$ &           0.13$^3$ &           0.18$^3$ &           0.46$^3$ \\
                              & Notes+SDOH &              0.75$^1$ &  0.26$^1$ &           0.05$^1$ &           0.01$^1$ &  0.06$^1$ &           0.56$^1$ &  0.09$^1$ &           0.37$^1$ &           0.32$^1$ &           0.38$^1$ &           0.72$^2$ &  0.51$^2$ &           0.13$^2$ &           0.18$^2$ &           0.49$^2$ \\
                              & All &              0.82$^3$ &  0.38$^3$ &           0.09$^3$ &           0.11$^3$ &  0.44$^3$ &           0.61$^3$ &  0.12$^3$ &           0.36$^3$ &           0.27$^3$ &           0.46$^3$ &           0.77$^3$ &  0.57$^3$ &           0.12$^3$ &           0.17$^3$ &           0.55$^3$ \\
                              & All+SDOH &              0.81$^2$ &  0.39$^2$ &            0.1$^2$ &           0.11$^2$ &  0.41$^2$ &           0.58$^3$ &  0.13$^3$ &   \textbf{0.3}$^3$ &           0.31$^3$ &           0.48$^3$ &           0.77$^2$ &  0.56$^2$ &           0.12$^2$ &           0.18$^2$ &           0.56$^2$ \\

 \hline
\multicolumn{17}{l}{$^1$ Trained on CHR, $^2$ Trained on SVI, $^3$ Trained on SDOHD}
\end{tabular}%
}
\end{table}
\end{landscape}
\end{document}